\pdfoutput=1
\documentclass[11pt]{article}

\usepackage{acl}

\usepackage{times}
\usepackage{latexsym}
\usepackage{inconsolata}
\usepackage{enumitem}
\usepackage{graphicx}
\usepackage{amsmath}
\usepackage{amssymb}
\usepackage{booktabs}
\usepackage{multirow}
\usepackage{colortbl}

\usepackage[T1]{fontenc}
\usepackage[utf8]{inputenc}

\usepackage{microtype}

\definecolor{fu-yellow}{RGB}{218,165,32} % yellow text

\newcommand{\std}[1]{\textsubscript{#1}}

\title{Meta-training with Demonstration Retrieval for\\ Efficient Few-shot Learning}

\author{Aaron Mueller$^{1}$\Thanks{ Work done as an intern at Meta.}\ \ \ \ \ Kanika Narang$^2$\ \ \ \ \ Lambert Mathias$^2$\\\textbf{Qifan Wang}$^2$\ \ \ \ \  \textbf{Hamed Firooz}$^2$\\
  $^1$ Johns Hopkins University, Baltimore, MD\\
  $^2$ Meta AI, Menlo Park, CA \\
  \texttt{amueller@jhu.edu}, \ \ \texttt{\{kanika13,mathiasl,wqfcr,mhfirooz\}@meta.com}\\}

\begin{document}
\maketitle
\begin{abstract}
Large language models show impressive results on few-shot NLP tasks. However, these models are memory and computation-intensive. Meta-training allows one to leverage smaller models for few-shot generalization in a domain-general and task-agnostic manner \citep{min2022metaicl,wei2022zeroshot,chen-etal-2022-meta}; however, these methods alone results in models that may not have sufficient parameterization or knowledge to adapt quickly to a large variety of tasks. To overcome this issue, we propose meta-training \emph{with demonstration retrieval}, where we use a dense passage retriever to retrieve semantically similar labeled demonstrations to each example for more varied supervision. By separating external knowledge from model parameters, we can use meta-training to train parameter-efficient models that generalize well on a larger variety of tasks. We construct a meta-training set from \textsc{UnifiedQA} and \textsc{CrossFit}, and propose a demonstration bank based on \textsc{UnifiedQA} tasks. To our knowledge, our work is the first to combine retrieval with meta-training, to use DPR models to retrieve demonstrations, and to leverage demonstrations from many tasks simultaneously, rather than randomly sampling demonstrations from the training set of the target task. Our approach outperforms a variety of targeted parameter-efficient and retrieval-augmented few-shot methods on QA, NLI, and text classification tasks (including SQuAD, QNLI, and TREC). Our approach can be meta-trained and fine-tuned quickly on a single GPU.
\end{abstract}

\section{Introduction}
\begin{figure}
    \centering
    \includegraphics[width=\linewidth]{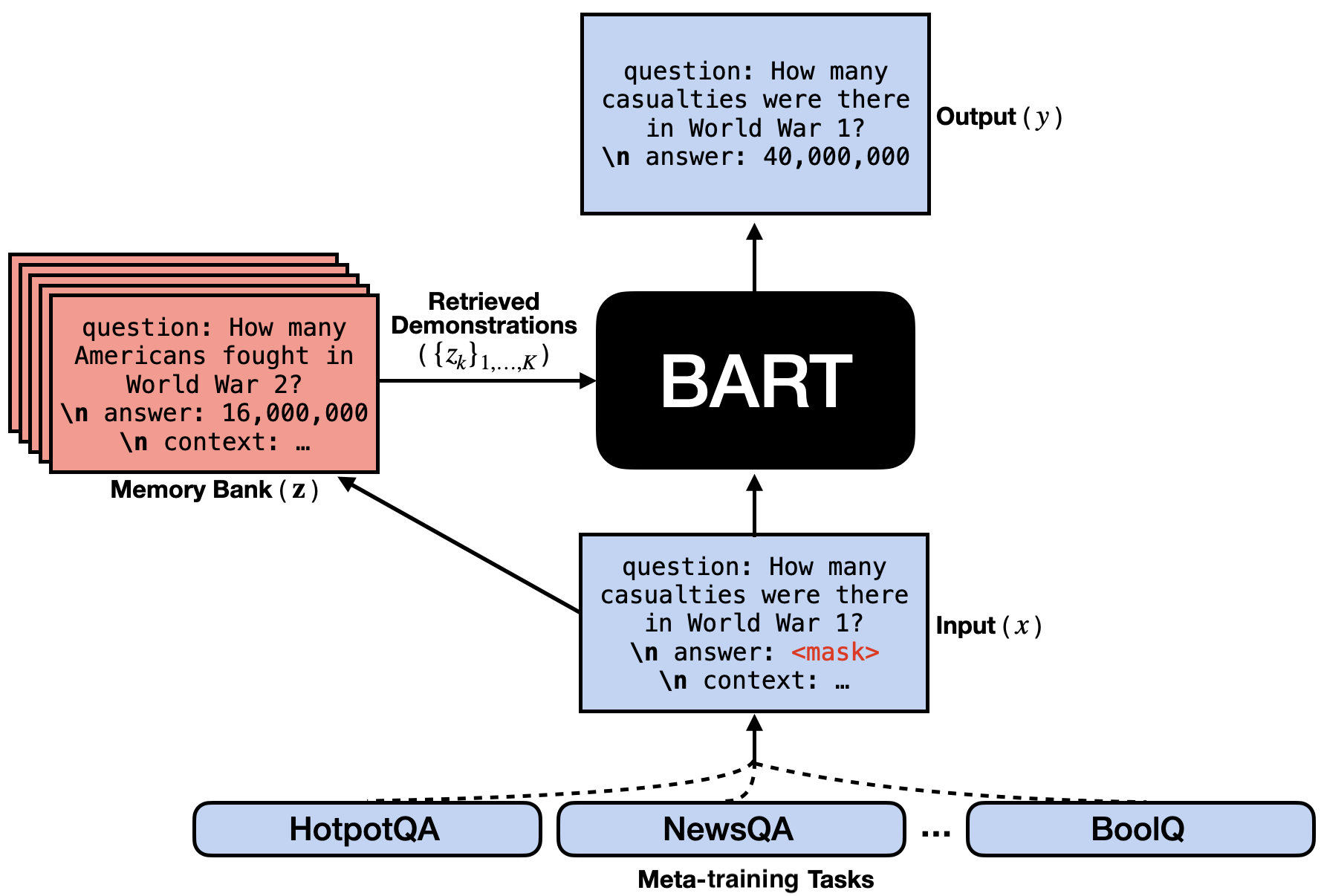}
    \caption{Our approach. Given an input $x$ from one of many possible QA tasks, we use a dense passage retriever to retrieve $K$ semantically similar demonstrations $Z=\{z_k\}_{1,\ldots,K}$ from a memory bank $\mathbf{z}$ composed of labeled examples. We meta-train BART, supervising it to generate the (question and) answer $y$ given $x$ and $Z$ across a diverse collection of QA tasks.}
    \label{fig:metalearn}
\end{figure}

Large language models (LLMs) have become increasingly popular due to their impressive few-shot performance on many NLP tasks and domains \citep{brown2020gpt3,chowdhery2022palm}. This has resulted in many few-shot learning methods based on LLMs that require ever-larger GPUs and increasing computation. Methods requiring no parameter updates such as in-context learning \citep{brown2020gpt3} and parameter-efficient methods like Adapters \citep{houlsby2019parameter} partially mitigate these downsides, but ultimately, larger computation budgets are increasingly necessary to achieve state-of-the-art few-shot performance---even to simply load models and perform inference.

Meta-learning \citep{vilalta2002metalearn,finn2017metalearn} and meta-training \citep{min2022metaicl} are methods that make smaller language models capable of quicker and more robust few-shot performance across multiple tasks and domains. However, smaller models may not be able to store enough knowledge for effective generalization in many domains and tasks simultaneously. Retrieval is one way to overcome this: by separating parametric knowledge in the language model from external knowledge (stored as retrievable text), one can leverage much more information than could be stored in the parameters of a language model. For example, retrieval-augmented generation (RAG; \citealp{lewis2020rag}) and retrieval-enhanced transformers (RETRO; \citealp{borgeaud2022improving}) retrieve natural language passages to improve performance on knowledge-intensive NLP tasks, although they do not perform meta-learning or meta-training and only evaluate on high-resource knowledge-intensive tasks.

We thus propose \textbf{meta-training with demonstration retrieval} as a more parameter-efficient way to leverage demonstrations for few-shot learning. We retrieve semantically similar labeled demonstrations for each training and test example during meta-training \emph{and} fine-tuning. On a relatively small sequence-to-sequence model (BART\textsubscript{large}, 440M parameters), we show our proposed approach is capable of generalizing quickly and well on a variety of downstream tasks (Table~\ref{tab:eval_data}). Inspired by retrieval-augmented generation (RAG) models \citep{lewis2020rag}, we use a dense passage retriever (DPR; \citealp{karpukhin2020dpr}) to retrieve demonstrations instead of Wikipedia passages. We retrieve semantically similar demonstrations from a large and diverse bank (\S\ref{sec:demo_memory}) that is compiled from many existing question answering tasks (App.\ \ref{app:tasks}), rather than randomly sampling demonstrations from the training set of the target task like most contemporary work \citep{min2022metaicl,brown2020gpt3,gao2021lmbff}.

Our experiments show that our method (\S\ref{sec:method}) outperforms tailored efficient few-shot baselines and other retrieval-augmented models on various tasks, including natural language inference (NLI), paraphrase detection, and extractive question answering (\S\ref{sec:results}). To our knowledge, our work is the first to combine retrieval with meta-training (or multi-task training more broadly), to use DPR models to retrieve demonstrations, and to leverage demonstrations from many tasks simultaneously, rather than retrieving random or $k$-nearest demonstrations from the training set of the target task.

Our code is available on GitHub.\footnote{\url{https://github.com/facebookresearch/metatrained-demRAG}}

\section{Related Work}
\textbf{Meta-learning} \citep{vilalta2002metalearn,finn2017metalearn} is a class of methods that supervise a model on \emph{how to learn}; the goal is to leverage a collection of meta-training tasks to learn a better learning algorithm that generalizes to held-out tasks. Inspired by meta-learning, some recent studies have attempted to induce specific abilities in language models in a task- and domain-agnostic manner via \textbf{meta-training}; this entails directly supervising a model on labeled examples from various tasks (sometimes using some controlled format or template \citep{chen-etal-2022-meta,wei2022zeroshot}) to directly induce specific abilities or better inductive biases that improve generalization. Meta-training is typically accomplished via some form of controlled multi-task learning, as in \citet{min2022metaicl}. Many studies have explored multi-task and multi-domain learning \citep{khashabi2020unifiedqa,zhong2021adapting,aghajanyan2021muppet,ye2021crossfit,wei2022zeroshot}, but these studies often leverage tasks that improve a model's abilities for some specific (set of) downstream tasks. In meta-training, we aim to directly improve the learning algorithm via controlled supervision, which should improve out-of-distribution generalization by teaching a model some helpful ability---such as in-context learning---that can result in gains on various downstream tasks \citep{min2022metaicl}.
We focus on meta-training with examples from QA datasets.

\textbf{Few-shot learning} is a common setting in which a model is supervised on only a few labeled examples. Many methods for improving few-shot performance are based on scaling model and data size \citep{brown2020gpt3,chowdhery2022palm}. Our goal is to improve few-shot performance across tasks in a computation- and memory-efficient manner, so we focus on smaller models that can be trained efficiently on a single GPU. Some parameter-efficient few-shot methods have been proposed, including cloze-style prompting \citep{schick2021pet}, fine-tuning with manually tuned \citep{schick2021exploiting} and automatically tuned prompts and demonstrations \citep{gao2021lmbff}, and meta-learning \citep{yu2018diverse,bansal2020self,bao2020fewshot}. One advantage of our approach is that it does not require significant prompt tuning: rather, we  standardize all of our tasks into a single format, similar to \citet{chada2021fewshotqa}. This saves human time and computational resources.

Crucially, these approaches compare probabilities of single tokens or small pre-selected label sets; thus, they cannot be used for open-domain tasks like question answering. Some work has proposed \emph{generative} few-shot methods for open-domain tasks: this includes reformatting the input data to match a model's pre-training format \citep{chada2021fewshotqa}, pre-training models to select relevant spans from context passages \citep{ram2021splinter}, and running a secondary pre-training step on labeled classification data \citep{mueller2022label}. Our model should be effective on many tasks, even when the label space is large and differs across examples; thus, our method is based on a \emph{generative} sequence-to-sequence model.

%\textbf{In-context learning}
\textbf{In-context learning} (ICL; \citealp{brown2020gpt3}) is increasingly used in few-shot methods; here, labeled \emph{demonstrations} are concatenated to the same context as a test example to teach a model how to perform a task without additional gradient updates. Studies have analyzed what kinds of demonstrations are most effective \citep{liu2022examples}, as well as what makes demonstrations effective \citep{min2022rethinking,xie2022bayesianicl}. Our demonstration retrieval approach is most similar to \citet{liu2022examples}, who encode demonstrations and test examples into a sentence embedding space and retrieve the $k$-nearest demonstrations. Our method differs in multiple ways: we use dense passage retrievers instead of sentence embeddings; we use demonstrations from many training sets instead of the training set of the target task; and we perform gradient updates with demonstrations, which is more feasible on our relatively small BART\textsubscript{large}-based model.

\citet{wei2022zeroshot} find that very large LMs ($>$68B parameters) are required for ICL to be effective, but \citet{min2022metaicl} find that meta-training can be used to make a much smaller model (GPT2\textsubscript{\text{large}}, 774M parameters) capable of leveraging demonstrations. Here, we make BART\textsubscript{large} (440M parameters) better at leveraging demonstrations through meta-training with demonstrations, like \citet{min2022metaicl}; however, their method is designed for zero-shot generalization, and it selects from a constrained set of pre-defined labels. Our method is designed for \emph{few-shot} settings and can be applied to open-domain tasks.

\textbf{Retrieval-augmented generation models} consist of two components: \emph{generators} and \emph{retrievers}. The generator is typically a decoder-only LM \citep{guu2020realm} or sequence-to-sequence (seq2seq) model \citep{lewis2020rag,izacard2020fid}; we use seq2seq models. The retriever is most often a dense passage retrieval (DPR; \citealp{karpukhin2020dpr}) model based on BERT\textsubscript{base}. RAG models are typically evaluated on knowledge-intensive tasks like abstractive QA and fact verification. Thus, the memory bank typically consists of Wikipedia passages, which augments the model with additional factual knowledge separate from the generator's parameters. \citet{izacard2022atlas} adapts this architecture for few-shot knowledge-intensive tasks using a very large generator (T5\textsubscript{X(X)L}) and a Contriever-based \citep{izacard2022contriever} retriever. However, we are interested in more general-purpose methods, as well as more parameter- and memory-efficient methods that train or fine-tune quickly on a single GPU. Thus, we propose a task-agnostic and domain-general method to improve smaller generative models for few-shot settings: specifically, a retrieval-augmented meta-training step and a memory bank of labeled QA demonstrations instead of Wikipedia passages.

\begin{table*}[t]
    \centering
    \resizebox{0.85\linewidth}{!}{
    \begin{tabular}{p{2.5cm}llrrrrr}
    \toprule
    \bf{Category} & \bf{Dataset} & \bf{Type} & \#Train & \#Test & $L$ & $|\mathcal{Y}|$\\
    \midrule
    \multirow{3}{*}{\parbox{2.5cm}{Extractive QA}} & SQuAD \citep{rajpurkar2016squad} & Open QA & 86,588 & 10,507 & 10 / 120 & - \\
    & BioASQ \citep{tsatsaronis2015bioasq} & Open QA & 24,559 & 1,504 & 10 / 200 & - \\
    & QASC \citep{khot2020qasc} & Multi-choice QA & 8,134 & 926 & 8 / 18 & - \\
    \midrule
    \multirow{2}{*}{\parbox{2.5cm}{Knowledge-intensive QA}} & TriviaQA \citep{joshi2017triviaqa} & Open QA & 61,688 & 7,785 & 13 / 677 & - \\
    & TextbookQA \citep{kembhavi2017textbookqa} & Open QA & 15,154 & 1,503 & 10 / 581 & - \\
    \midrule
    \multirow{5}{*}{\parbox{2.5cm}{Classification}} & TREC \citep{voorhees2000trec} & Question class.\ & 5,452 & 500 & 10 & 6 \\
    & MRPC \citep{dolan2005mrpc} & Paraphrase class.\ & 3,668 & 408 & 22 / 21 & 2 \\
    & MNLI \citep{williams2018mnli} & NLI & 392,702 & 9,815 & 22 / 11 & 3 \\
    & MNLI-mm (\emph{ibid.}) & NLI & 392,702 & 9,832 & 22 / 11 & 3 \\
    & QNLI \citep{wang2018glue} & NLI & 104,743 & 5,463 & 11 / 30 & 3 \\
    \bottomrule
    \end{tabular}}
    \caption{Evaluation sets used in this study. $L$: mean \# of words in question/context or input sentence(s). For more straightforward comparison to prior few-shot question answering and classification methods, we use \citet{ram2021splinter}'s few-shot splits of SQuAD and BioASQ derived from MRQA, %\footnote{\url{https://github.com/mrqa/MRQA-Shared-Task-2019}}
    as well as \citet{gao2021lmbff}'s splits of TREC, MRPC, MNLI(-mm), and QNLI. We generate our own few-shot splits of QASC using 5 random seeds for each split size.}
    \label{tab:eval_data}
\end{table*}

\section{Method}\label{sec:method}
\subsection{Retrieval-augmented Generation}\label{sec:rag}
As we wish to retrieve similar labeled examples for every input, our architecture takes inspiration from retrieval-augmented generation (RAG) models \citep{lewis2020rag}, which consist of a pre-trained sequence-to-sequence component (we use BART\textsubscript{large}) and a pre-trained dense passage retriever (DPR) component. Given an input $x$, the DPR component retrieves the $K$ most semantically similar memory entries $\{z_k\}_{1,\ldots,K}$ from the memory bank $\mathbf{z}$. Retrieval is performed using a BERT-based input encoder $E_I$ on $x$ and BERT-based demonstration encoder $E_D$ on $\mathbf{z}$ to encode both into a vector space, and then running maximum inner product search:\footnote{Maximum inner product search can be approximately solved in sub-linear time \citep{johnson2019billion}. We use the \texttt{faiss} library for this: \url{https://github.com/facebookresearch/faiss}}
\begin{align}
    \{z_k\}_{1,\ldots,K} = \underset{z\in\mathbf{z}}{\text{top-}K}\left\{ E_I(x)^\top E_D(z)) \right\}
\end{align}
The DPR component also returns the inner products themselves as document scores $p_\eta(z_k|x)$.

The input and retrieved entries are then passed to a pre-trained sequence-to-sequence model, BART\textsubscript{large}, for autoregressive generation. At each timestep, we marginalize over the retrieved demonstrations by creating $K$ separate input contexts, consisting of the input $x$ and one retrieved entry $z_k$. We then sum over BART's token probabilities $p_\theta$ given each context, weighted by $z_k$'s document score:\footnote{This is similar to the RAG-Token approach in \citet{lewis2020rag}. The number of demonstrations we can use is \emph{not} limited by the context length since we marginalize over each demonstration in its own separate context.}
\begin{align}
    p(y|x) \approx \prod_i^N \sum_k^K p_\eta(z_k|x) p_\theta(y|x,z_k,y_{1:i-1})
\end{align}

\subsection{Meta-training}\label{sec:meta_learning}
To adapt a sequence-to-sequence model for general-purpose demonstration retrieval and answer generation, we perform a meta-training step by supervising the model with demonstrations on a collection\footnote{Throughout this study, we use ``task'' to refer to a single dataset like SQuAD or NaturalQuestions, and ``collection'' to refer to the dataset obtained by concatenating a set of tasks.} of 18 QA tasks (Table~\ref{tab:metatrain_tasks}). We update the parameters of the BART component of our model during meta-training by supervising BART (using its normal cross-entropy loss) to generate the question and its answer given the question and a set of retrieved demonstrations. We use QA tasks due to the semantic diversity of inputs and labels; compare to text classification tasks, where the label space is much smaller and labels are often less informative.

We modify and use the QA meta-training task collections from \citep{min2022metaicl}.% The non-QA set consists of 30+ text classification tasks from \textsc{CrossFit} \citep{ye2021crossfit}, including MNLI, AG News, GLUE tasks, \emph{inter alia}; the QA set 
This consists of various extractive, multiple-choice, and/or abstractive QA tasks from \textsc{CrossFit} \emph{and} a subsample of \textsc{UnifiedQA} \citep{khashabi2020unifiedqa,khashabi2022unifiedqa}, including NaturalQuestions, MCTest, BIOMRC, \emph{inter alia}. We modify the meta-training collections by (1) removing our evaluation sets if they are present,\footnote{We also remove any examples where the question has a Jaccard similarity $>0.9$ with \emph{any} training or test question in our evaluation tasks, \emph{and} where the answers are the same; only 4 such examples existed in our data.} and (2) standardizing the format of each task. Our final meta-training collection contains 32 tasks, which we subsample to 18 tasks based on semantic similarity to our evaluation tasks; see Appendix~\ref{app:tasks} for a full list of tasks and details on our semantic subsampling procedure, and \S\ref{sec:ablations} for a description of the downstream effect of semantic subsampling.

Following \citet{chada2021fewshotqa}, we standardize each input in the meta-training data to a ``\texttt{question:}\ldots\ \texttt{\textbackslash{}n} \texttt{answer:}\ \texttt{[MASK]}\ \texttt{\textbackslash{}n} \texttt{context:}\ldots'' format. Then, the output sequence consists of both the question \emph{and} answer sequences,\footnote{We only compute $F_1$ on the answer sequences.} which aligns with BART's pre-training objective of reconstructing the entire input sequence (not just masked spans). Like \citet{chada2021fewshotqa}, we find that aligning the input/output format with BART's pre-training objective makes a positive difference for downstream performance. For QASC, which is a multiple-choice QA task, we put all of the answer options in the \texttt{context} field before the two context sentences and generate the full answer string. This outperformed all other formats we tried by a significant margin.\footnote{We tried placing the answer options in the \texttt{question} field, not including the answer options at all, and only generating the letter label instead of the full answer string. See Appendix~\ref{app:qasc_formats} for examples and scores.}

For classification tasks, we use the same question/answer/context format. For our single-sentence classification task (TREC), we place the input in the \texttt{question} field, and present all of the possible labels in the \texttt{context} field using a similar format as for QASC. For sentence-pair classification tasks (MRPC, MNLI(-mm), QNLI), we place the first sentence or hypothesis in the \texttt{question} field and place the second sentence or premise in the \texttt{context} field. As with QA tasks, we generate both the question and answer fields in the target sequence, but only evaluate $F_1$ on answer sequences.

\subsection{Demonstration Memory}\label{sec:demo_memory}
For the demonstration memory bank, we use training sets from \textsc{UnifiedQA}, excluding our evaluation tasks; the memory contains examples from 16 tasks. UnifiedQA has approximately 40\% overlap with the QA meta-training collection, and no overlap with the non-QA collection. See Table~\ref{tab:demo_tasks} in Appendix~\ref{app:tasks} for a full list of tasks in our demonstration memory bank.

We format each demonstration in the memory bank in the same question/answer/context format as described above, except that demonstrations have the ground-truth label after the \texttt{answer:} header instead of a \texttt{[MASK]} token. Note that memory entries consist of a text passage (the demonstration) \emph{and} a title; for the title, we simply use the answer to the question.

\begin{figure*}
    \centering
    \includegraphics[width=0.56\linewidth]{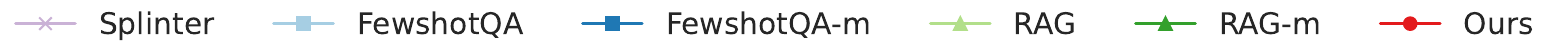}
    
    \includegraphics[width=0.32\linewidth]{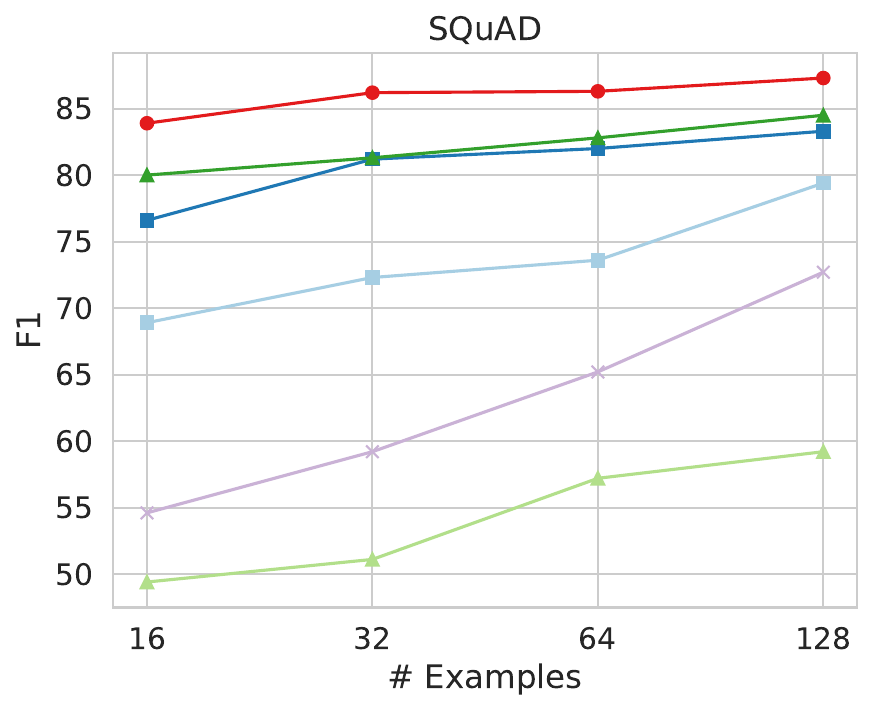}
    \includegraphics[width=0.32\linewidth]{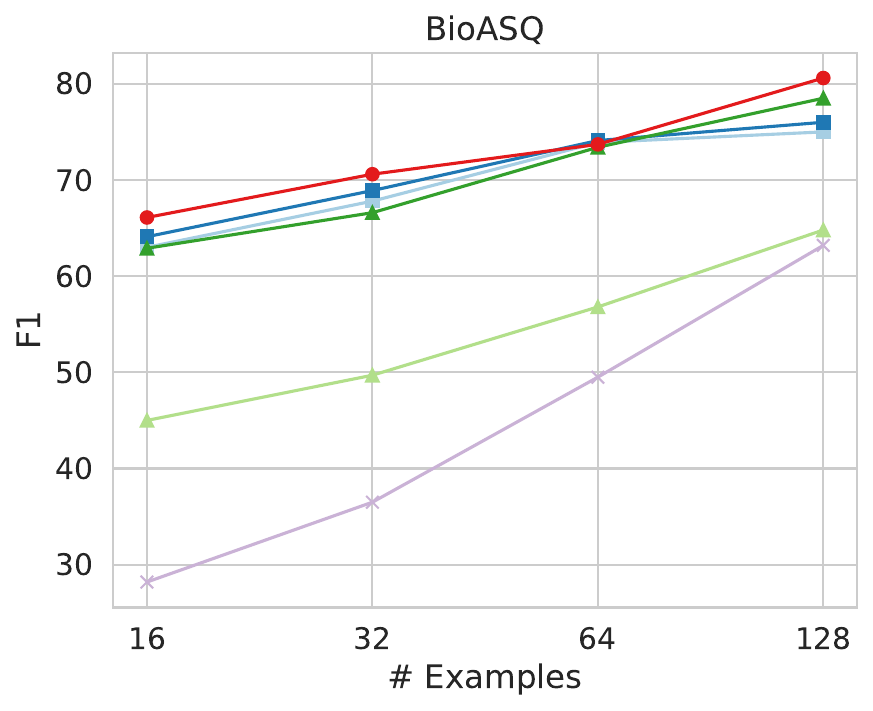}
    \includegraphics[width=0.32\linewidth]{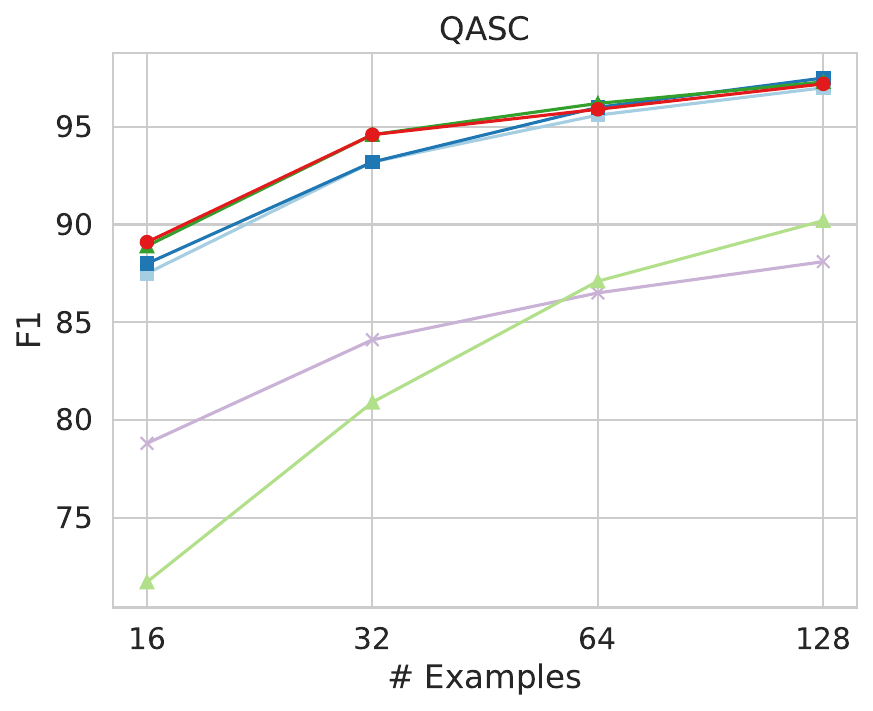}
    \caption{$F_1$ scores at each few-shot split size for extractive and multiple-choice question answering evaluation tasks. Scores are averaged across 5 random few-shot samples. Our model outperforms or maintains similar performance to the strongest baselines on each task and split size. Performance gains on SQuAD are especially large---up to 3.9 $F_1$ (4.9\% improvement). FewshotQA and Splinter scores are from \citet{chada2021fewshotqa}.}
    \label{fig:qa}
\end{figure*}

\section{Experimental Setup}\label{sec:experiments}
We evaluate on a variety of QA and classification tasks (Table~\ref{tab:eval_data}). We select open-domain QA tasks from the MRQA shared task \citep{fisch2019mrqa} to reflect a variety of extractive QA formats, including a standard QA benchmark (SQuAD), a domain-specific challenging benchmark (BioASQ), and two knowledge-intensive QA benchmarks (TriviaQA and TextbookQA).\footnote{While ``knowledge-intensive'' does not have a standard definition or straightforward measurement, the length of the contexts may act as a proxy for how knowledge-intensive a question answering task is. Contexts for our knowledge-intensive tasks are much longer, and thus require a model to synthesize much more information and/or retrieve information that is more relevant to the inputs to semantically prime the model for question-specific information.} Our few-shot QA splits of size \{16, 32, 64, 128\} for these tasks are from \citet{ram2021splinter}, which are themselves derived from MRQA \citep{fisch2019mrqa}. We also generate few-shot splits for QASC, which is a multiple-choice QA task; we evaluate on QASC to determine whether our model is also effective in dealing with much shorter contexts, and to ensure that it is not overfitting to more typical MRQA-style extractive tasks.

Our few-shot classification task splits are from \citet{gao2021lmbff}. We evaluate on sentence pair classification tasks which are not contained in our meta-training or demonstration tasks; sentence pair classification tasks like natural language inference (NLI) and paraphrase classification can be easily reformatted to our question/answer/context format. We also evaluate on TREC, which is a single-sentence text classification task where the model must guess the \emph{category} of the answer to a question (e.g., human, location, number), rather than the answer itself.

For each task and few-shot split size, we average scores across 5 random few-shot samples.

\subsection{Baselines}
We compare against strong efficient few-shot methods, as well as similar models that will tell us \emph{why} our method performs better. Note that our approach is \emph{generative}, unlike iPET and LM-BFF; thus, it is usable on a wider variety of tasks.

\textbf{FewshotQA} \citep{chada2021fewshotqa}. A few-shot question answering method. We compare to the FewshotBARTL model, which is based on BART\textsubscript{large} like our model and is the best-performing variant. We use the same few-shot splits such that we can directly compare to the numbers reported in that paper. We also try meta-training this non-retrieval-augmented model, which is essentially our method without retrieval; we call this baseline FewshotQA-m.

\textbf{Splinter} \citep{ram2021splinter}. A few-shot question answering model pre-trained to select salient spans from context passages.

\textbf{RAG} \citep{lewis2020rag}. The original RAG-Token model with a memory of Wikipedia passages. We use the released model fine-tuned on NaturalQuestions (NQ), as this was the best-performing RAG model on our tasks. To see whether our demonstration memory is more effective than Wikipedia passages when meta-training, we also try meta-training the RAG model with its Wikipedia memory; we call this baseline RAG-m.

\textbf{iPET} \citep{schick2021pet}. A manual prompt-tuning approach that induces better few-shot performance than GPT3 with much smaller LMs. We tune the best-performing ALBERT\textsubscript{xxl} \citep{lan2020albert} model on our tasks.

\textbf{LM-BFF} \citep{gao2021lmbff}. An automatic prompt-tuning approach based on RoBERTa\textsubscript{large} \citep{liu2019roberta}. It requires no unlabeled text data to work well, unlike iPET. This model and iPET compare token probabilities to perform classification, so we cannot use them for open-domain tasks like question answering. Thus, we only compare to these models on classification.

\subsection{Hyperparameters}
For meta-training, we use hyperparameters from \citet{min2022metaicl} where possible: init.\ LR 1$\times$10$^{-5}$, effective batch size 8,\footnote{We use gradient accumulation to get this effective batch size on a single GPU.} training for a maximum of 30,000 steps. We checkpoint every 2,000 steps and select the checkpoint with the lowest mean loss on our 16-shot QA training sets. Meta-training finishes in $\approx$14 hours on 1 A100 GPU (40GB).\footnote{Our model could also be trained and tuned on a cheaper 32GB GPU (e.g., a V100) in similar time.}

For fine-tuning, we use hyperparameters from \citet{chada2021fewshotqa} where possible: init.\ LR $2\times 10^{-5}$, batch size $4$, fine-tuning for a maximum of 1,000 steps or 35 epochs (whichever is larger). We checkpoint every 2 epochs and select the checkpoint with the highest exact match on the training set. Fine-tuning finishes in 30--60 minutes on 1 A100 GPU (40GB).

For each meta-training and fine-tuning input, we retrieve $5$ demonstrations from the memory.\footnote{Higher values result in better performance (\S\ref{sec:ablations}), but this saturates at 5--10 retrieved demonstrations, and retrieving more demonstrations slows down training.}

\begin{table}[]
    \centering
    \resizebox{\linewidth}{!}{
        \begin{tabular}{lrrrrrr}
            \toprule
            %Model & SQuAD & BioASQ & QASC & TriviaQA & TbQA \\
            %\midrule
            %FewshotQA & & & & \\
            %RAG & & & & \\
            %FiD & & & & \\
            %Ours & & & & \\
            %\midrule \midrule
            & TREC & MNLI & MNLI-mm & QNLI & MRPC & \emph{Avg.} \\
            \midrule
            Majority & 18.8 & 32.7 & 33.3 & 49.5 & 81.2 & \emph{43.1} \\
            \midrule
            RoBERTa & *88.8\std{2.1} & *45.8\std{6.4} & *47.8\std{6.8} & *60.2\std{6.5} & 76.6\std{2.5} & \emph{63.8} \\
            iPET & *85.0\std{4.1} & 71.2\std{1.7} & 71.8\std{2.6} & *70.3\std{6.2} & 70.4\std{4.7} & \emph{73.7} \\
            LM-BFF & *89.4\std{1.7} & 70.7\std{1.3} & *\textbf{72.0}\std{1.2} & *69.2\std{1.9} & *\textbf{78.1}\std{3.4} & \emph{75.9} \\
            FewshotQA & 91.0\std{2.0} & *47.9\std{6.3} & *46.1\std{5.9} & *61.0\std{6.4} & *67.6\std{4.8} & \emph{62.7} \\
            FewshotQA-m & \textbf{92.4}\std{1.4} & *50.1\std{1.0} & *50.6\std{2.5} & *71.8\std{2.1} & 74.0\std{3.7} & \emph{67.8} \\
            RAG & *81.1\std{2.0} & *62.4\std{0.9} & *61.8\std{1.2} & *74.9\std{1.5} & 70.2\std{3.3} & \emph{70.1} \\
            RAG-m & *87.8\std{1.7} & *70.0\std{1.4} & 69.1\std{1.4} & *83.2\std{1.5} & 74.9\std{2.8} & \emph{77.0} \\
            \midrule
            Ours & 91.7\std{1.3} & \textbf{72.9}\std{1.7} & 69.6\std{1.4} & \textbf{84.4}\std{1.8} & 73.4\std{2.5} & \textbf{\emph{78.4}} \\
            \bottomrule
        \end{tabular}
    }
    \caption{Accuracies on classification tasks, averaged across 5 random few-shot samples (std.\ dev.\ in subscript). All datasets are well-balanced except MRPC; thus, we report accuracies for all tasks except MRPC, where we report macro-$F_1$. LM-BFF and RoBERTa scores are from \citet{gao2021lmbff}. * indicates that $p<.05$ in a $t$-test between our model's score and the marked score.}
    \label{tab:classification}
\end{table}

\section{Results}\label{sec:results}
Our model's $F_1$ scores for extractive question answering (Figure~\ref{fig:qa}) are higher than models of similar parameterizations, including similar models that have been meta-trained using the same training data. Our model also outperforms strong classification approaches on TREC, MNLI, and QNLI (Table~\ref{tab:classification}). Thus, \textbf{meta-training with semantically similar demonstrations induces a more general-purpose system that can perform well across a variety of low-resource downstream tasks.}

Contrast this with RAG, which often performs \emph{worst} out of each model we test across tasks. Thus, the architecture itself is not inherently strong in few-shot settings, suggesting that meta-training makes a significant contribution to increased performance. This is also supported by the increased performance we observe with FewshotQA and RAG after meta-training, though note that meta-training does not help FewshotQA to the same extent it helps retrieval-augmented models. Also note that FewshotQA does not perform well on classification tasks, whereas our method achieves performance exceeding or close to the strongest baselines. This means that the combination of meta-training and retrieval enables a more general-purpose model than either of these components separately.

With meta-training, RAG-m obtains performance much closer to our model. This tells us that \textbf{meta-training is responsible for much of the performance gains we observe}, though the demonstration memory bank also improves performance to a lesser extent. On MRPC, RAG-m outperforms our model, indicating that there exist some non-knowledge-intensive tasks where Wikipedia passages are more helpful than QA demonstrations.

\subsection{Knowledge-intensive QA}\label{sec:knowledge_intensive_results}
\begin{figure}[t]
    \centering
    \includegraphics[width=\linewidth]{results_figs/legend.pdf}
    
    \includegraphics[width=0.75\linewidth]{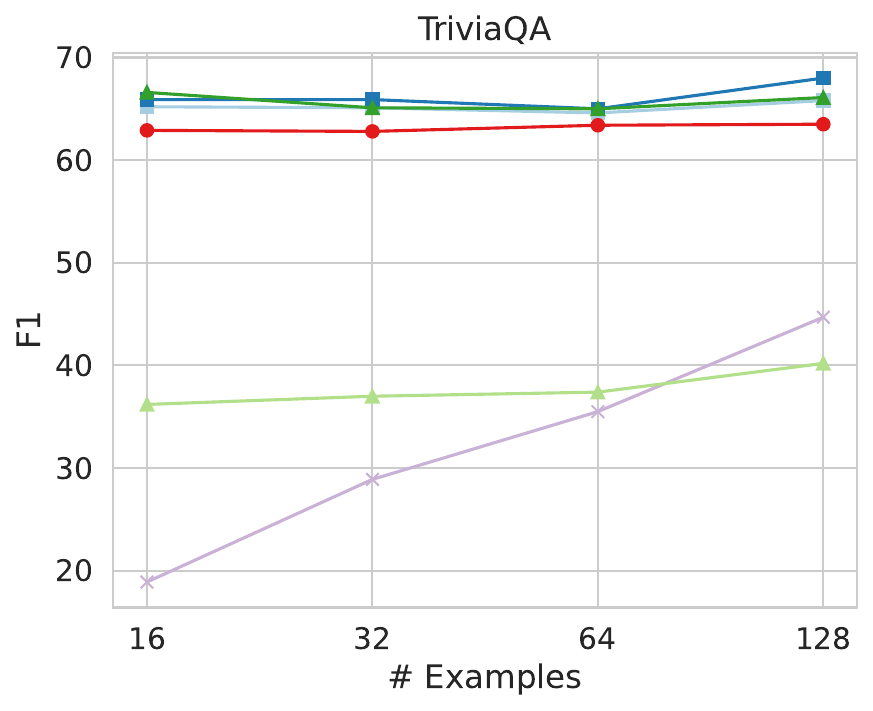}
    \includegraphics[width=0.75\linewidth]{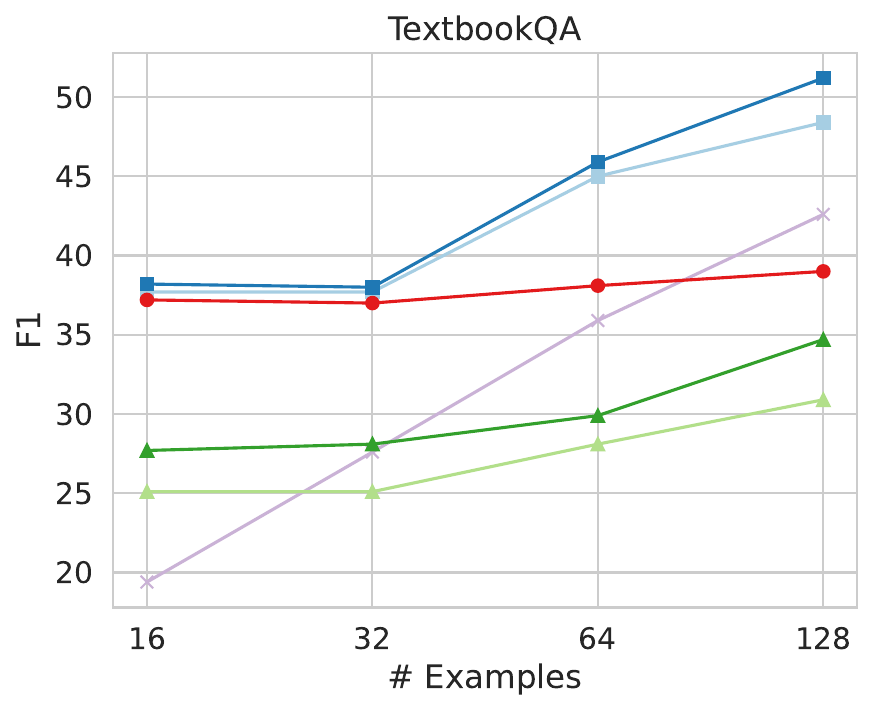}
    \caption{$F_1$ scores for each few-shot split size for knowledge-intensive question answering tasks. Our model is outperformed by a strong few-shot QA baseline, though meta-training still greatly improves performance.}
    \label{fig:knowledge_intensive}
\end{figure}

We also evaluate on few-shot knowledge-intensive QA tasks (Figure~\ref{fig:knowledge_intensive}): here, TriviaQA and TextbookQA, using the few-shot splits from the MRQA shared task. While these are also technically extractive QA tasks, their contexts have an average length of 677 and 581 words, respectively, meaning that BART will likely struggle more to synthesize all of the information in these tasks (even with retrieval). We find that FewshotQA outperforms our method on both of these tasks, and that even Splinter outperforms our method at larger split sizes for TextbookQA. This means that demonstration retrieval may be actively harmful for these tasks. Thus, our meta-training method is optimizing RAG architectures for non-knowledge-intensive tasks, but not for knowledge-intensive tasks. Wikipedia passages are more effective than demonstrations in the memory bank for TriviaQA as well, as indicated by RAG-m outperforming our approach.

However, meta-training with or without the memory bank still induces far better performance than the base RAG model, which performs worse than all baselines except Splinter. Thus, our method is still improving over RAG, making this model more versatile and better able to handle such tasks even if it is not the optimal approach.

\subsection{Ablations}\label{sec:ablations}
Here, we perform further analyses to understand the contribution of individual model components and (meta-)training decisions.

\textbf{Memory bank.} We find that performance is generally higher for question answering and classification when retrieving demonstrations instead of Wikipedia passages, as in Figure~\ref{fig:qa} and Table~\ref{tab:classification}. This raises two questions: how much could the memory bank impact downstream performance in the best-case scenario? Relatedly, what is the upper bound on performance for our model given the best possible demonstration memory bank?

To obtain an estimate, we create an \emph{oracle} memory consisting of labeled test examples from our evaluation data. We find that scores significantly improve over our method and others in this setting, indicating that \textbf{this architecture has significant potential to achieve further gains if the memory bank is improved}.

\begin{table}[t]
    \centering
    \resizebox{\linewidth}{!}{
    \begin{tabular}{lrrrrr}
    \toprule
    \textbf{Model} & SQuAD & BioASQ & QASC & TriviaQA & TbQA \\
    \midrule
    FewshotQA & 68.9 & 63.0 & 82.6 & 65.2 & 37.7 \\
    FewshotQA-m & 76.6 & 63.4 & 85.9 & 65.9 & \bf{38.2} \\
    RAG-m & 80.0 & 62.9 & 88.9 & \bf{66.6} & 27.7 \\
    Ours & \bf{83.9} & \bf{64.7} & \bf{89.2} & 62.9 & 37.2 \\
    \midrule
    Ours (\emph{oracle}) & 93.5 & 94.2 & 99.1 & 80.7 & 83.2 \\
    \bottomrule
    \end{tabular}}
    \caption{$F_1$ scores on QA tasks for our strongest baselines, our approach, and our approach where the memory has been replaced with labeled test examples (\emph{oracle}). The oracle approach establishes an approximate upper bound for our model. Large gaps between our approach and the oracle indicate room for improvement in what constitutes our memory bank.}
    \label{tab:oracle_memory}
\end{table}

\begin{figure}
    \centering
    \includegraphics[width=0.48\linewidth]{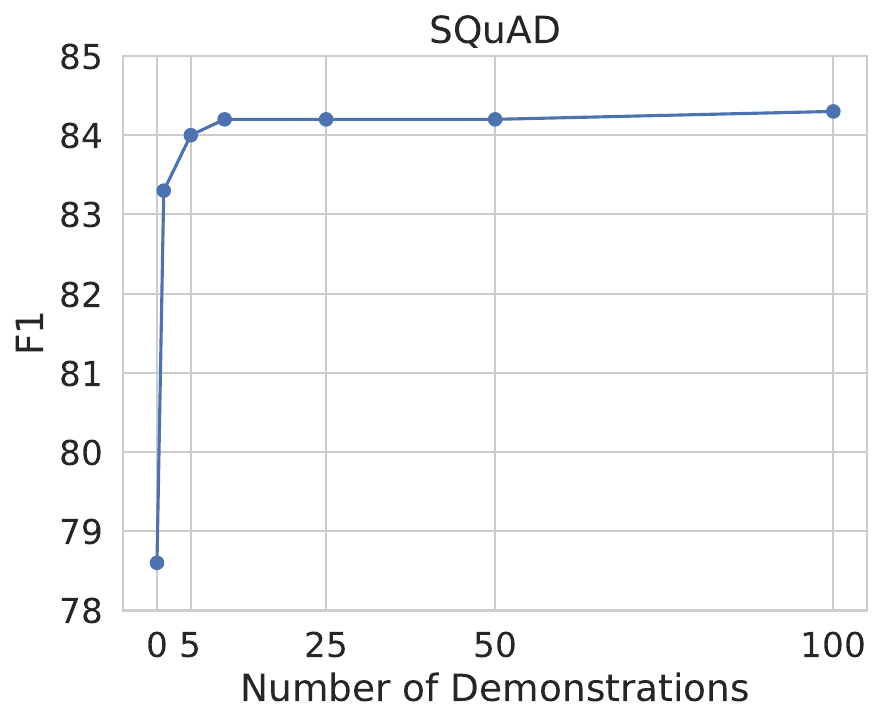}
    \includegraphics[width=0.48\linewidth]{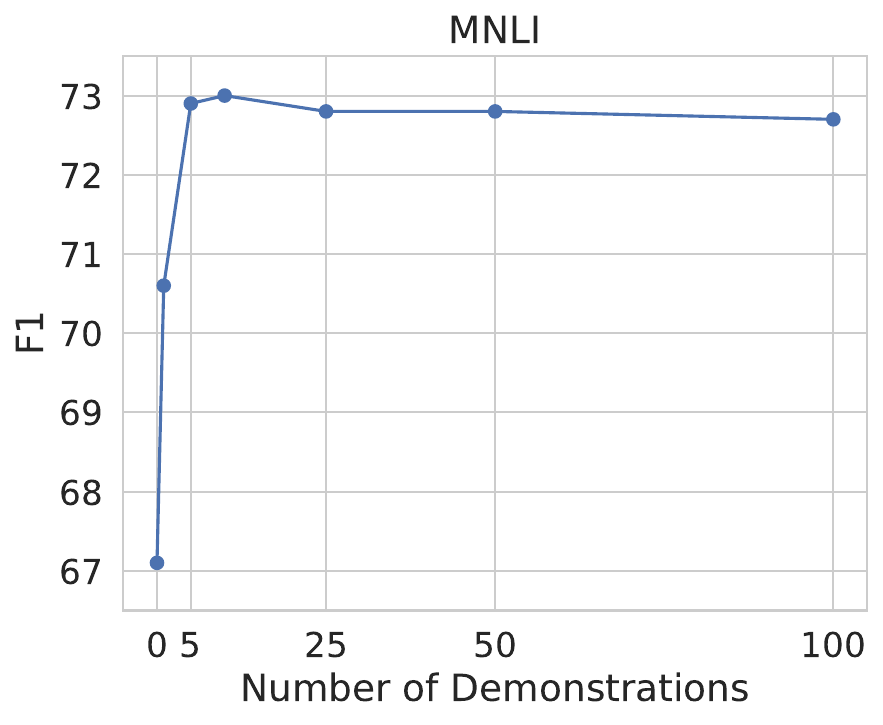}
    \caption{$F_1$ scores for an extractive QA (SQuAD) and sentence pair classification (MNLI) task by the number of retrieved demonstrations (\{$0,1,5,10,25,50,100$\}) during fine-tuning. Scores generally increase with the number of retrieved demonstrations, though performance saturates early at $5$--$10$ demonstrations.}
    \label{fig:num_docs}
\end{figure}

\textbf{Number of retrieved demonstrations.} Is retrieving more demonstrations always better? We compare performance when retrieving $K=$~\{0,~1,~5,~10,~25,~50\} demonstrations during fine-tuning and evaluation on non-knowledge-intensive QA (SQuAD) and sentence-pair classification (MNLI). Our results (Figure~\ref{fig:num_docs}) show that $F_1$ scores begin to saturate at 5--10 demonstrations for both tasks. However, using more demonstrations generally does not harm performance; the model is able to handle less helpful demonstrations without performance decreasing significantly.

\begin{figure}
    \centering
    \includegraphics[width=0.48\linewidth]{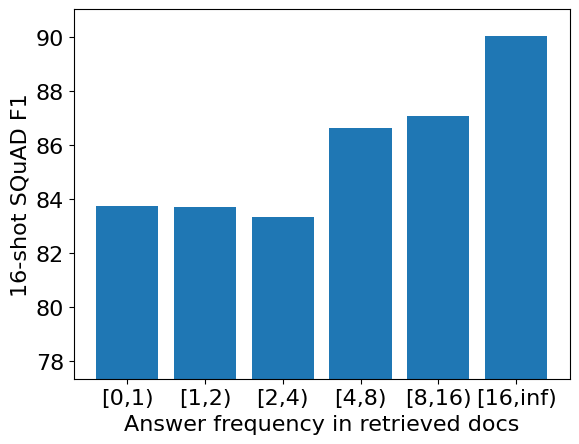}
    \includegraphics[width=0.48\linewidth]{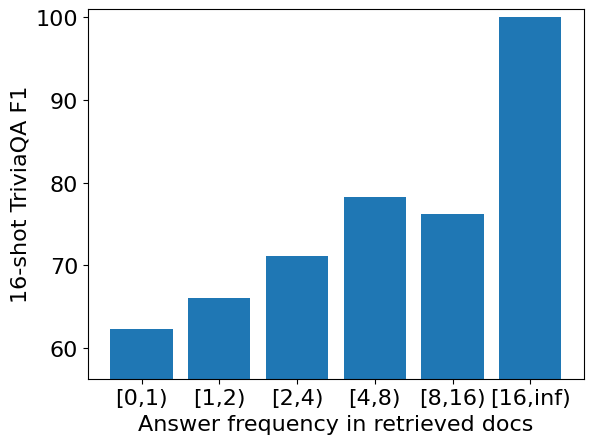}
    \caption{$F_1$ scores for non-knowledge-intensive (SQuAD, left) and knowledge-intensive (TriviaQA, right) QA tasks by the frequency of the true answer string in the retrieved demonstrations. While not monotonic, there is a clear correlation between these variables, indicating that lexical features may be responsible for much of retrieval's contributions to performance.}
    \label{fig:passage_overlap}
\end{figure}

\textbf{Why is retrieval helpful?} Is the model abstracting semantic content from the retrieved demonstrations for improved performance, or is it simply learning to copy token sequences from the retrieved demonstrations? As an initial test, we can correlate the frequency of the ground-truth answer sequence in the retrieved documents with $F_1$ scores on our QA tasks. Our results (Figure~\ref{fig:passage_overlap}) suggest that the model is indeed learning to retrieve certain text strings from the demonstrations. This provides one possible path forward for improving the memory bank: higher semantic overlap with one's evaluation task increases the likelihood of these overlaps, so future work could focus on collecting (or perhaps generating) more semantically similar demonstrations that feature more lexical overlaps.

\begin{table}[t]
    \centering
    \resizebox{\linewidth}{!}{
    \begin{tabular}{lrrrrr}
        \toprule
        \textbf{Retriever} & SQuAD & BioASQ & QASC & TriviaQA & TbQA \\
        \midrule
        \emph{Random} & \emph{1.8} & \emph{1.5} & \emph{1.2} & \emph{1.8} & \emph{2.3} \\
        DPR (Wiki) & 11.5 & 1.8 & 15.7 & 4.9 & \textbf{24.3} \\
        DPR (PAQ) & \textbf{16.9} & 1.5 & 26.1 & \textbf{29.3} & 24.0 \\
        Contriever & 14.1 & \textbf{7.3} & \textbf{28.0} & 27.9 & \textbf{24.3} \\
        \bottomrule
    \end{tabular}}
    \caption{The proportion of test examples for which each retriever retrieves at least 1 demonstration containing the ground-truth answer as a substring. DPR (PAQ) and Contriever appear to be better at retrieving more relevant demonstrations on average, though this does not necessarily lead to higher downstream performance (Table~\ref{tab:app_retriever_downstream}).}
    \label{tab:app_retriever_freq}
\end{table}

\begin{table}[]
    \centering
    \resizebox{\linewidth}{!}{
    \begin{tabular}{lrrrrr}
    \toprule
    \textbf{Retriever} & SQuAD & BioASQ & QASC & TriviaQA & TbQA \\
    \midrule
    \emph{Random} & \emph{74.3} & \emph{61.8} & \emph{88.7} & \emph{56.5} & \emph{29.6} \\
    DPR (Wiki) & \textbf{83.9} & \textbf{64.7} & \textbf{89.2} & \textbf{62.9} & \textbf{37.2} \\
    DPR (PAQ) & 78.8 & 63.5 & 86.8 & 57.6 & 33.5 \\
    Contriever & 81.1 & 62.5 & 88.7 & 58.9 & 32.4 \\
    \bottomrule
    \end{tabular}}
    \caption{$F_1$ scores on 16-shot extractive QA tasks across retrievers. We fine-tune with different retrievers given the same (best) meta-trained model. Despite DPR (Wiki)'s lower retriever scores (Table~\ref{tab:app_retriever_freq}), its downstream performance is the best among the retrievers we try.}
    \label{tab:app_retriever_downstream}
\end{table}

However, this does not explain how retrieval improves performance on classification tasks, where the label space is small and labels are less informative. For NLI, the label space includes ``entailment''/``neutral''/``contradiction'', which we would not expect to see often in our demonstrations and which do not carry significant semantic content. Yet retrieval-augmented models outperform FewshotQA by a large margin on MNLI(-mm), so what is helping our model? There could exist some QA demonstrations which semantically prime our model toward correct completions, though sentence embedding similarity may not capture this helpfulness. Future work could ablate over specific features in the demonstrations.

\paragraph{What type of retriever is best?} For our experiments thus far, we have used the DPR component of the RAG-Token (NQ) model, which is pre-trained on Wikipedia and fine-tuned on NaturalQuestions. Is this an optimal starting point, or would some other retreiver be better? We compare against a DPR model pre-trained on the Probably-Asked Questions (PAQ; \citealp{lewis2021paq}) dataset, as well as the Contriever model \citep{izacard2022contriever}. Contrievers are unsupervised, whereas DPR models receive explicit supervision during pre-training. DPR tends to perform better when the downstream task is similar to the pre-training or fine-tuning data; however, in our case, demonstration retrieval is dissimilar from Wikipedia passage retrieval, and Contriever may handle larger train-test shifts better \citep{izacard2022contriever}.

We evaluate both the relevance of the retrieved demonstrations (Table~\ref{tab:app_retriever_freq}) and downstream $F_1$ (Table~\ref{tab:app_retriever_downstream}) on our QA tasks. We find that DPR (PAQ) and Contriever are both better at retrieving similar demonstrations, as measured by the frequency with which they retrieve examples that contain the answer. For BioASQ, only Contriever retrieves more relevant demonstrations than a random retriever.

However, retrieving more relevant demonstrations does not translate into increased downstream performance: DPR (Wiki) consistently outperforms the others. Why? Through qualitative analysis, we find that DPR (Wiki) retrieves more semantically diverse demonstrations, whereas DPR (PAQ) and Contriever retrieve demonstrations that are technically more similar to the test example, but also less diverse \emph{across} test examples.%(thus decreasing the diversity of supervision received during fine-tuning).
Thus, there should be a balance between diversity and relevance: completely random retrieval is not effective (as indicated by our random retrieval baseline scoring worst), but neither is the more constrained demonstration set we retrieve using an arguably more optimal retriever.

\begin{table}[]
    \centering
    \resizebox{\linewidth}{!}{
    \begin{tabular}{lrrrrr}
    \toprule
    \textbf{Memory} & SQuAD & BioASQ & QASC & TriviaQA & TbQA \\
    \midrule
    All tasks & 83.5 & 63.2 & \bf{89.2} & 61.4 & 36.8 \\
    Semantically similar tasks & \bf{83.9} & \bf{64.7} & \bf{89.2} & \bf{62.9} & \bf{37.2} \\
    \bottomrule
    \end{tabular}}
    \caption{16-shot $F_1$ scores on QA tasks after meta-training on either all QA tasks from MetaICL's QA meta-training collection, or QA tasks subsampled by semantic similarity to our evaluation tasks. A full list of meta-training tasks can be found in Appendix~\ref{app:tasks}.}
    \label{tab:metalearn_data}
\end{table}

\textbf{Meta-training data.} Is meta-training helpful because of the variety of tasks included in our setup (the \emph{more is better} hypothesis), or would it be better to select meta-training data in a more principled way (the \emph{similar datasets are better} hypothesis)? We compare downstream performance when meta-training on all QA tasks from MetaICL versus the top tasks by mean instance-level semantic similarity to our evaluation tasks (Table~\ref{tab:metalearn_data}). To compute semantic similarity, we use the 
\texttt{stsb-roberta-base-v2} model from SentenceTransformers \citep{reimers2019sentence} and compute the mean pairwise cosine similarity between the 16-shot training examples in our evaluation tasks and all examples in a meta-training task. We then select the top tasks by similarity until we have over 240,000 examples (enough for 30,000 training steps using batch size 8). See Appendix~\ref{app:tasks} for a list of meta-training tasks before and after subsampling.

We find that \textbf{selecting meta-training data based on semantic similarity to our evaluation tasks is helpful for both our QA \emph{and} non-QA tasks}: $F_1$ increases across tasks when only meta-training on the most similar data. This contrasts with the findings of \citet{min2022metaicl}, who find that more meta-training tasks is generally better.

\section{Conclusions}
We have proposed a meta-training method (\S\ref{sec:meta_learning}) that retrieves (\S\ref{sec:rag}) semantically similar demonstrations from a diverse demonstration bank (\S\ref{sec:demo_memory}). Our method achieves higher performance on average across many tasks than other strong parameter-efficient few-shot baselines (\S\ref{sec:results}). In future work, one could explore a mixture of demonstration retrieval and passage retrieval for improved performance on a wider variety of tasks---including knowledge-intensive tasks.

\section*{Limitations}
Our method requires access to a large set of labeled examples for the memory bank---ideally with some relevance to the evaluation tasks. This limits the languages and tasks that are optimal for this method: there does not exist a large variety of training examples for low-resource language varieties, nor for certain much more specific tasks---as in, for example, industry applications with domain-specific customer data. And while multilingual models could leverage cross-lingual transfer, it is unclear how well this model would generalize into low-resource languages when (for example) using multilingual BART. 

When using the full demonstration memory, meta-training does not run on a 16GB GPU using our current implementation. While this does exclude more common GPUs, our approach could still run quickly on a 32GB GPU in a few hours, thus costing far less than pre-training a language model of comparable few-shot performance from scratch.

\bibliography{anthology,custom}

\begin{thebibliography}{47}
\expandafter\ifx\csname natexlab\endcsname\relax\def\natexlab#1{#1}\fi

\bibitem[{Aghajanyan et~al.(2021)Aghajanyan, Gupta, Shrivastava, Chen,
  Zettlemoyer, and Gupta}]{aghajanyan2021muppet}
Armen Aghajanyan, Anchit Gupta, Akshat Shrivastava, Xilun Chen, Luke
  Zettlemoyer, and Sonal Gupta. 2021.
\newblock \href {https://doi.org/10.18653/v1/2021.emnlp-main.468} {Muppet:
  Massive multi-task representations with pre-finetuning}.
\newblock In \emph{Proceedings of the 2021 Conference on Empirical Methods in
  Natural Language Processing}, pages 5799--5811, Online and Punta Cana,
  Dominican Republic. Association for Computational Linguistics.

\bibitem[{Bansal et~al.(2020)Bansal, Jha, Munkhdalai, and
  McCallum}]{bansal2020self}
Trapit Bansal, Rishikesh Jha, Tsendsuren Munkhdalai, and Andrew McCallum. 2020.
\newblock \href {https://doi.org/10.18653/v1/2020.emnlp-main.38}
  {Self-supervised meta-learning for few-shot natural language classification
  tasks}.
\newblock In \emph{Proceedings of the 2020 Conference on Empirical Methods in
  Natural Language Processing (EMNLP)}, pages 522--534, Online. Association for
  Computational Linguistics.

\bibitem[{Bao et~al.(2020)Bao, Wu, Chang, and Barzilay}]{bao2020fewshot}
Yujia Bao, Menghua Wu, Shiyu Chang, and Regina Barzilay. 2020.
\newblock \href {https://openreview.net/forum?id=H1emfT4twB} {Few-shot text
  classification with distributional signatures}.
\newblock In \emph{8th International Conference on Learning Representations,
  {ICLR} 2020, Addis Ababa, Ethiopia, April 26-30, 2020}. OpenReview.net.

\bibitem[{Borgeaud et~al.(2022)Borgeaud, Mensch, Hoffmann, Cai, Rutherford,
  Millican, van~den Driessche, Lespiau, Damoc, Clark, de~Las~Casas, Guy,
  Menick, Ring, Hennigan, Huang, Maggiore, Jones, Cassirer, Brock, Paganini,
  Irving, Vinyals, Osindero, Simonyan, Rae, Elsen, and
  Sifre}]{borgeaud2022improving}
Sebastian Borgeaud, Arthur Mensch, Jordan Hoffmann, Trevor Cai, Eliza
  Rutherford, Katie Millican, George van~den Driessche, Jean-Baptiste Lespiau,
  Bogdan Damoc, Aidan Clark, Diego de~Las~Casas, Aurelia Guy, Jacob Menick,
  Roman Ring, Tom Hennigan, Saffron Huang, Loren Maggiore, Chris Jones, Albin
  Cassirer, Andy Brock, Michela Paganini, Geoffrey Irving, Oriol Vinyals, Simon
  Osindero, Karen Simonyan, Jack~W. Rae, Erich Elsen, and Laurent Sifre. 2022.
\newblock \href {http://arxiv.org/abs/2112.04426} {Improving language models by
  retrieving from trillions of tokens}.

\bibitem[{Brown et~al.(2020)Brown, Mann, Ryder, Subbiah, Kaplan, Dhariwal,
  Neelakantan, Shyam, Sastry, Askell, Agarwal, Herbert-Voss, Krueger, Henighan,
  Child, Ramesh, Ziegler, Wu, Winter, Hesse, Chen, Sigler, Litwin, Gray, Chess,
  Clark, Berner, McCandlish, Radford, Sutskever, and Amodei}]{brown2020gpt3}
Tom Brown, Benjamin Mann, Nick Ryder, Melanie Subbiah, Jared~D Kaplan, Prafulla
  Dhariwal, Arvind Neelakantan, Pranav Shyam, Girish Sastry, Amanda Askell,
  Sandhini Agarwal, Ariel Herbert-Voss, Gretchen Krueger, Tom Henighan, Rewon
  Child, Aditya Ramesh, Daniel Ziegler, Jeffrey Wu, Clemens Winter, Chris
  Hesse, Mark Chen, Eric Sigler, Mateusz Litwin, Scott Gray, Benjamin Chess,
  Jack Clark, Christopher Berner, Sam McCandlish, Alec Radford, Ilya Sutskever,
  and Dario Amodei. 2020.
\newblock \href
  {https://proceedings.neurips.cc/paper/2020/file/1457c0d6bfcb4967418bfb8ac142f64a-Paper.pdf}
  {Language models are few-shot learners}.
\newblock In \emph{Advances in Neural Information Processing Systems},
  volume~33, pages 1877--1901. Curran Associates, Inc.

\bibitem[{Chada and Natarajan(2021)}]{chada2021fewshotqa}
Rakesh Chada and Pradeep Natarajan. 2021.
\newblock \href {https://doi.org/10.18653/v1/2021.emnlp-main.491}
  {{F}ewshot{QA}: A simple framework for few-shot learning of question
  answering tasks using pre-trained text-to-text models}.
\newblock In \emph{Proceedings of the 2021 Conference on Empirical Methods in
  Natural Language Processing}, pages 6081--6090, Online and Punta Cana,
  Dominican Republic. Association for Computational Linguistics.

\bibitem[{Chen et~al.(2022)Chen, Zhong, Zha, Karypis, and
  He}]{chen-etal-2022-meta}
Yanda Chen, Ruiqi Zhong, Sheng Zha, George Karypis, and He~He. 2022.
\newblock \href {https://doi.org/10.18653/v1/2022.acl-long.53} {Meta-learning
  via language model in-context tuning}.
\newblock In \emph{Proceedings of the 60th Annual Meeting of the Association
  for Computational Linguistics (Volume 1: Long Papers)}, pages 719--730,
  Dublin, Ireland. Association for Computational Linguistics.

\bibitem[{Chowdhery et~al.(2022)Chowdhery, Narang, Devlin, Bosma, Mishra,
  Roberts, Barham, Chung, Sutton, Gehrmann, Schuh, Shi, Tsvyashchenko, Maynez,
  Rao, Barnes, Tay, Shazeer, Prabhakaran, Reif, Du, Hutchinson, Pope, Bradbury,
  Austin, Isard, Gur{-}Ari, Yin, Duke, Levskaya, Ghemawat, Dev, Michalewski,
  Garcia, Misra, Robinson, Fedus, Zhou, Ippolito, Luan, Lim, Zoph, Spiridonov,
  Sepassi, Dohan, Agrawal, Omernick, Dai, Pillai, Pellat, Lewkowycz, Moreira,
  Child, Polozov, Lee, Zhou, Wang, Saeta, Diaz, Firat, Catasta, Wei,
  Meier{-}Hellstern, Eck, Dean, Petrov, and Fiedel}]{chowdhery2022palm}
Aakanksha Chowdhery, Sharan Narang, Jacob Devlin, Maarten Bosma, Gaurav Mishra,
  Adam Roberts, Paul Barham, Hyung~Won Chung, Charles Sutton, Sebastian
  Gehrmann, Parker Schuh, Kensen Shi, Sasha Tsvyashchenko, Joshua Maynez,
  Abhishek Rao, Parker Barnes, Yi~Tay, Noam Shazeer, Vinodkumar Prabhakaran,
  Emily Reif, Nan Du, Ben Hutchinson, Reiner Pope, James Bradbury, Jacob
  Austin, Michael Isard, Guy Gur{-}Ari, Pengcheng Yin, Toju Duke, Anselm
  Levskaya, Sanjay Ghemawat, Sunipa Dev, Henryk Michalewski, Xavier Garcia,
  Vedant Misra, Kevin Robinson, Liam Fedus, Denny Zhou, Daphne Ippolito, David
  Luan, Hyeontaek Lim, Barret Zoph, Alexander Spiridonov, Ryan Sepassi, David
  Dohan, Shivani Agrawal, Mark Omernick, Andrew~M. Dai,
  Thanumalayan~Sankaranarayana Pillai, Marie Pellat, Aitor Lewkowycz, Erica
  Moreira, Rewon Child, Oleksandr Polozov, Katherine Lee, Zongwei Zhou, Xuezhi
  Wang, Brennan Saeta, Mark Diaz, Orhan Firat, Michele Catasta, Jason Wei,
  Kathy Meier{-}Hellstern, Douglas Eck, Jeff Dean, Slav Petrov, and Noah
  Fiedel. 2022.
\newblock \href {https://doi.org/10.48550/arXiv.2204.02311} {Palm: Scaling
  language modeling with pathways}.
\newblock \emph{CoRR}, abs/2204.02311.

\bibitem[{Dolan and Brockett(2005)}]{dolan2005mrpc}
William~B. Dolan and Chris Brockett. 2005.
\newblock \href {https://aclanthology.org/I05-5002} {Automatically constructing
  a corpus of sentential paraphrases}.
\newblock In \emph{Proceedings of the Third International Workshop on
  Paraphrasing ({IWP}2005)}.

\bibitem[{Finn et~al.(2017)Finn, Abbeel, and Levine}]{finn2017metalearn}
Chelsea Finn, Pieter Abbeel, and Sergey Levine. 2017.
\newblock \href {http://proceedings.mlr.press/v70/finn17a.html} {Model-agnostic
  meta-learning for fast adaptation of deep networks}.
\newblock In \emph{Proceedings of the 34th International Conference on Machine
  Learning, {ICML} 2017, Sydney, NSW, Australia, 6-11 August 2017}, volume~70
  of \emph{Proceedings of Machine Learning Research}, pages 1126--1135. {PMLR}.

\bibitem[{Fisch et~al.(2019)Fisch, Talmor, Jia, Seo, Choi, and
  Chen}]{fisch2019mrqa}
Adam Fisch, Alon Talmor, Robin Jia, Minjoon Seo, Eunsol Choi, and Danqi Chen.
  2019.
\newblock \href {https://doi.org/10.18653/v1/D19-5801} {{MRQA} 2019 shared
  task: Evaluating generalization in reading comprehension}.
\newblock In \emph{Proceedings of the 2nd Workshop on Machine Reading for
  Question Answering, MRQA@EMNLP 2019, Hong Kong, China, November 4, 2019},
  pages 1--13. Association for Computational Linguistics.

\bibitem[{Gao et~al.(2021)Gao, Fisch, and Chen}]{gao2021lmbff}
Tianyu Gao, Adam Fisch, and Danqi Chen. 2021.
\newblock \href {https://doi.org/10.18653/v1/2021.acl-long.295} {Making
  pre-trained language models better few-shot learners}.
\newblock In \emph{Proceedings of the 59th Annual Meeting of the Association
  for Computational Linguistics and the 11th International Joint Conference on
  Natural Language Processing (Volume 1: Long Papers)}, pages 3816--3830,
  Online. Association for Computational Linguistics.

\bibitem[{Guu et~al.(2020)Guu, Lee, Tung, Pasupat, and Chang}]{guu2020realm}
Kelvin Guu, Kenton Lee, Zora Tung, Panupong Pasupat, and Ming{-}Wei Chang.
  2020.
\newblock \href {http://proceedings.mlr.press/v119/guu20a.html} {Retrieval
  augmented language model pre-training}.
\newblock In \emph{Proceedings of the 37th International Conference on Machine
  Learning, {ICML} 2020, 13-18 July 2020, Virtual Event}, volume 119 of
  \emph{Proceedings of Machine Learning Research}, pages 3929--3938. {PMLR}.

\bibitem[{Houlsby et~al.(2019)Houlsby, Giurgiu, Jastrzebski, Morrone,
  de~Laroussilhe, Gesmundo, Attariyan, and Gelly}]{houlsby2019parameter}
Neil Houlsby, Andrei Giurgiu, Stanislaw Jastrzebski, Bruna Morrone, Quentin
  de~Laroussilhe, Andrea Gesmundo, Mona Attariyan, and Sylvain Gelly. 2019.
\newblock \href {http://proceedings.mlr.press/v97/houlsby19a.html}
  {Parameter-efficient transfer learning for {NLP}}.
\newblock In \emph{Proceedings of the 36th International Conference on Machine
  Learning, {ICML} 2019, 9-15 June 2019, Long Beach, California, {USA}},
  volume~97 of \emph{Proceedings of Machine Learning Research}, pages
  2790--2799. {PMLR}.

\bibitem[{Izacard et~al.(2021)Izacard, Caron, Hosseini, Riedel, Bojanowski,
  Joulin, and Grave}]{izacard2022contriever}
Gautier Izacard, Mathilde Caron, Lucas Hosseini, Sebastian Riedel, Piotr
  Bojanowski, Armand Joulin, and Edouard Grave. 2021.
\newblock \href {http://arxiv.org/abs/2112.09118} {Towards unsupervised dense
  information retrieval with contrastive learning}.
\newblock \emph{CoRR}, abs/2112.09118.

\bibitem[{Izacard and Grave(2021)}]{izacard2020fid}
Gautier Izacard and Edouard Grave. 2021.
\newblock \href {https://doi.org/10.18653/v1/2021.eacl-main.74} {Leveraging
  passage retrieval with generative models for open domain question answering}.
\newblock In \emph{Proceedings of the 16th Conference of the European Chapter
  of the Association for Computational Linguistics: Main Volume, {EACL} 2021,
  Online, April 19 - 23, 2021}, pages 874--880. Association for Computational
  Linguistics.

\bibitem[{Izacard et~al.(2022)Izacard, Lewis, Lomeli, Hosseini, Petroni,
  Schick, Dwivedi{-}Yu, Joulin, Riedel, and Grave}]{izacard2022atlas}
Gautier Izacard, Patrick Lewis, Maria Lomeli, Lucas Hosseini, Fabio Petroni,
  Timo Schick, Jane Dwivedi{-}Yu, Armand Joulin, Sebastian Riedel, and Edouard
  Grave. 2022.
\newblock \href {https://doi.org/10.48550/arXiv.2208.03299} {Few-shot learning
  with retrieval augmented language models}.
\newblock \emph{CoRR}, abs/2208.03299.

\bibitem[{Johnson et~al.(2021)Johnson, Douze, and
  J{\'{e}}gou}]{johnson2019billion}
Jeff Johnson, Matthijs Douze, and Herv{\'{e}} J{\'{e}}gou. 2021.
\newblock \href {https://doi.org/10.1109/TBDATA.2019.2921572} {Billion-scale
  similarity search with gpus}.
\newblock \emph{{IEEE} Trans. Big Data}, 7(3):535--547.

\bibitem[{Joshi et~al.(2017)Joshi, Choi, Weld, and
  Zettlemoyer}]{joshi2017triviaqa}
Mandar Joshi, Eunsol Choi, Daniel Weld, and Luke Zettlemoyer. 2017.
\newblock \href {https://doi.org/10.18653/v1/P17-1147} {{T}rivia{QA}: A large
  scale distantly supervised challenge dataset for reading comprehension}.
\newblock In \emph{Proceedings of the 55th Annual Meeting of the Association
  for Computational Linguistics (Volume 1: Long Papers)}, pages 1601--1611,
  Vancouver, Canada. Association for Computational Linguistics.

\bibitem[{Karpukhin et~al.(2020)Karpukhin, Oguz, Min, Lewis, Wu, Edunov, Chen,
  and Yih}]{karpukhin2020dpr}
Vladimir Karpukhin, Barlas Oguz, Sewon Min, Patrick Lewis, Ledell Wu, Sergey
  Edunov, Danqi Chen, and Wen-tau Yih. 2020.
\newblock \href {https://doi.org/10.18653/v1/2020.emnlp-main.550} {Dense
  passage retrieval for open-domain question answering}.
\newblock In \emph{Proceedings of the 2020 Conference on Empirical Methods in
  Natural Language Processing (EMNLP)}, pages 6769--6781, Online. Association
  for Computational Linguistics.

\bibitem[{Kembhavi et~al.(2017)Kembhavi, Seo, Schwenk, Choi, Farhadi, and
  Hajishirzi}]{kembhavi2017textbookqa}
Aniruddha Kembhavi, Minjoon Seo, Dustin Schwenk, Jonghyun Choi, Ali Farhadi,
  and Hannaneh Hajishirzi. 2017.
\newblock \href {https://doi.org/10.1109/CVPR.2017.571} {Are you smarter than a
  sixth grader? textbook question answering for multimodal machine
  comprehension}.
\newblock In \emph{2017 IEEE Conference on Computer Vision and Pattern
  Recognition (CVPR)}, pages 5376--5384.

\bibitem[{Khashabi et~al.(2022)Khashabi, Kordi, and
  Hajishirzi}]{khashabi2022unifiedqa}
Daniel Khashabi, Yeganeh Kordi, and Hannaneh Hajishirzi. 2022.
\newblock \href {http://arxiv.org/abs/2202.12359} {Unifiedqa-v2: Stronger
  generalization via broader cross-format training}.
\newblock \emph{CoRR}, abs/2202.12359.

\bibitem[{Khashabi et~al.(2020)Khashabi, Min, Khot, Sabharwal, Tafjord, Clark,
  and Hajishirzi}]{khashabi2020unifiedqa}
Daniel Khashabi, Sewon Min, Tushar Khot, Ashish Sabharwal, Oyvind Tafjord,
  Peter Clark, and Hannaneh Hajishirzi. 2020.
\newblock \href {https://doi.org/10.18653/v1/2020.findings-emnlp.171}
  {Unifiedqa: Crossing format boundaries with a single {QA} system}.
\newblock In \emph{Findings of the Association for Computational Linguistics:
  {EMNLP} 2020, Online Event, 16-20 November 2020}, volume {EMNLP} 2020 of
  \emph{Findings of {ACL}}, pages 1896--1907. Association for Computational
  Linguistics.

\bibitem[{Khot et~al.(2020)Khot, Clark, Guerquin, Jansen, and
  Sabharwal}]{khot2020qasc}
Tushar Khot, Peter Clark, Michal Guerquin, Peter Jansen, and Ashish Sabharwal.
  2020.
\newblock \href {https://doi.org/10.1609/aaai.v34i05.6319} {Qasc: A dataset for
  question answering via sentence composition}.
\newblock \emph{Proceedings of the AAAI Conference on Artificial Intelligence},
  34(05):8082--8090.

\bibitem[{Lan et~al.(2020)Lan, Chen, Goodman, Gimpel, Sharma, and
  Soricut}]{lan2020albert}
Zhenzhong Lan, Mingda Chen, Sebastian Goodman, Kevin Gimpel, Piyush Sharma, and
  Radu Soricut. 2020.
\newblock \href {https://openreview.net/forum?id=H1eA7AEtvS} {{ALBERT:} {A}
  lite {BERT} for self-supervised learning of language representations}.
\newblock In \emph{8th International Conference on Learning Representations,
  {ICLR} 2020, Addis Ababa, Ethiopia, April 26-30, 2020}. OpenReview.net.

\bibitem[{Lewis et~al.(2020)Lewis, Perez, Piktus, Petroni, Karpukhin, Goyal,
  K\"{u}ttler, Lewis, Yih, Rockt\"{a}schel, Riedel, and Kiela}]{lewis2020rag}
Patrick Lewis, Ethan Perez, Aleksandra Piktus, Fabio Petroni, Vladimir
  Karpukhin, Naman Goyal, Heinrich K\"{u}ttler, Mike Lewis, Wen-tau Yih, Tim
  Rockt\"{a}schel, Sebastian Riedel, and Douwe Kiela. 2020.
\newblock \href
  {https://proceedings.neurips.cc/paper/2020/file/6b493230205f780e1bc26945df7481e5-Paper.pdf}
  {Retrieval-augmented generation for knowledge-intensive nlp tasks}.
\newblock In \emph{Advances in Neural Information Processing Systems},
  volume~33, pages 9459--9474. Curran Associates, Inc.

\bibitem[{Lewis et~al.(2021)Lewis, Wu, Liu, Minervini, K{\"u}ttler, Piktus,
  Stenetorp, and Riedel}]{lewis2021paq}
Patrick Lewis, Yuxiang Wu, Linqing Liu, Pasquale Minervini, Heinrich
  K{\"u}ttler, Aleksandra Piktus, Pontus Stenetorp, and Sebastian Riedel. 2021.
\newblock \href {https://doi.org/10.1162/tacl_a_00415} {{PAQ}: 65 million
  probably-asked questions and what you can do with them}.
\newblock \emph{Transactions of the Association for Computational Linguistics},
  9:1098--1115.

\bibitem[{Liu et~al.(2022)Liu, Shen, Zhang, Dolan, Carin, and
  Chen}]{liu2022examples}
Jiachang Liu, Dinghan Shen, Yizhe Zhang, Bill Dolan, Lawrence Carin, and Weizhu
  Chen. 2022.
\newblock \href {https://doi.org/10.18653/v1/2022.deelio-1.10} {What makes good
  in-context examples for {GPT}-3?}
\newblock In \emph{Proceedings of Deep Learning Inside Out (DeeLIO 2022): The
  3rd Workshop on Knowledge Extraction and Integration for Deep Learning
  Architectures}, pages 100--114, Dublin, Ireland and Online. Association for
  Computational Linguistics.

\bibitem[{Liu et~al.(2019)Liu, Ott, Goyal, Du, Joshi, Chen, Levy, Lewis,
  Zettlemoyer, and Stoyanov}]{liu2019roberta}
Yinhan Liu, Myle Ott, Naman Goyal, Jingfei Du, Mandar Joshi, Danqi Chen, Omer
  Levy, Mike Lewis, Luke Zettlemoyer, and Veselin Stoyanov. 2019.
\newblock \href {http://arxiv.org/abs/1907.11692} {Roberta: {A} robustly
  optimized {BERT} pretraining approach}.
\newblock \emph{CoRR}, abs/1907.11692.

\bibitem[{Min et~al.(2022{\natexlab{a}})Min, Lewis, Zettlemoyer, and
  Hajishirzi}]{min2022metaicl}
Sewon Min, Mike Lewis, Luke Zettlemoyer, and Hannaneh Hajishirzi.
  2022{\natexlab{a}}.
\newblock \href {https://doi.org/10.18653/v1/2022.naacl-main.201} {{M}eta{ICL}:
  Learning to learn in context}.
\newblock In \emph{Proceedings of the 2022 Conference of the North American
  Chapter of the Association for Computational Linguistics: Human Language
  Technologies}, pages 2791--2809, Seattle, United States. Association for
  Computational Linguistics.

\bibitem[{Min et~al.(2022{\natexlab{b}})Min, Lyu, Holtzman, Artetxe, Lewis,
  Hajishirzi, and Zettlemoyer}]{min2022rethinking}
Sewon Min, Xinxi Lyu, Ari Holtzman, Mikel Artetxe, Mike Lewis, Hannaneh
  Hajishirzi, and Luke Zettlemoyer. 2022{\natexlab{b}}.
\newblock \href {http://arxiv.org/abs/2202.12837} {Rethinking the role of
  demonstrations: What makes in-context learning work?}
\newblock \emph{CoRR}, abs/2202.12837.

\bibitem[{Mueller et~al.(2022)Mueller, Krone, Romeo, Mansour, Mansimov, Zhang,
  and Roth}]{mueller2022label}
Aaron Mueller, Jason Krone, Salvatore Romeo, Saab Mansour, Elman Mansimov,
  Yi~Zhang, and Dan Roth. 2022.
\newblock \href {https://doi.org/10.18653/v1/2022.acl-long.570} {Label semantic
  aware pre-training for few-shot text classification}.
\newblock In \emph{Proceedings of the 60th Annual Meeting of the Association
  for Computational Linguistics (Volume 1: Long Papers)}, pages 8318--8334,
  Dublin, Ireland. Association for Computational Linguistics.

\bibitem[{Rajpurkar et~al.(2016)Rajpurkar, Zhang, Lopyrev, and
  Liang}]{rajpurkar2016squad}
Pranav Rajpurkar, Jian Zhang, Konstantin Lopyrev, and Percy Liang. 2016.
\newblock \href {https://doi.org/10.18653/v1/D16-1264} {{SQ}u{AD}: 100,000+
  questions for machine comprehension of text}.
\newblock In \emph{Proceedings of the 2016 Conference on Empirical Methods in
  Natural Language Processing}, pages 2383--2392, Austin, Texas. Association
  for Computational Linguistics.

\bibitem[{Ram et~al.(2021)Ram, Kirstain, Berant, Globerson, and
  Levy}]{ram2021splinter}
Ori Ram, Yuval Kirstain, Jonathan Berant, Amir Globerson, and Omer Levy. 2021.
\newblock \href {https://aclanthology.org/2021.acl-long.239} {Few-shot question
  answering by pretraining span selection}.
\newblock In \emph{Proceedings of the 59th Annual Meeting of the Association
  for Computational Linguistics and the 11th International Joint Conference on
  Natural Language Processing (Volume 1: Long Papers)}, pages 3066--3079,
  Online. Association for Computational Linguistics.

\bibitem[{Reimers and Gurevych(2019)}]{reimers2019sentence}
Nils Reimers and Iryna Gurevych. 2019.
\newblock \href {https://doi.org/10.18653/v1/D19-1410} {Sentence-{BERT}:
  Sentence embeddings using {S}iamese {BERT}-networks}.
\newblock In \emph{Proceedings of the 2019 Conference on Empirical Methods in
  Natural Language Processing and the 9th International Joint Conference on
  Natural Language Processing (EMNLP-IJCNLP)}, pages 3982--3992, Hong Kong,
  China. Association for Computational Linguistics.

\bibitem[{Schick and Sch{\"u}tze(2021{\natexlab{a}})}]{schick2021exploiting}
Timo Schick and Hinrich Sch{\"u}tze. 2021{\natexlab{a}}.
\newblock \href {https://doi.org/10.18653/v1/2021.eacl-main.20} {Exploiting
  cloze-questions for few-shot text classification and natural language
  inference}.
\newblock In \emph{Proceedings of the 16th Conference of the European Chapter
  of the Association for Computational Linguistics: Main Volume}, pages
  255--269, Online. Association for Computational Linguistics.

\bibitem[{Schick and Sch{\"u}tze(2021{\natexlab{b}})}]{schick2021pet}
Timo Schick and Hinrich Sch{\"u}tze. 2021{\natexlab{b}}.
\newblock \href {https://doi.org/10.18653/v1/2021.naacl-main.185} {It{'}s not
  just size that matters: Small language models are also few-shot learners}.
\newblock In \emph{Proceedings of the 2021 Conference of the North American
  Chapter of the Association for Computational Linguistics: Human Language
  Technologies}, pages 2339--2352, Online. Association for Computational
  Linguistics.

\bibitem[{Tsatsaronis et~al.(2015)Tsatsaronis, Balikas, Malakasiotis, Partalas,
  Zschunke, Alvers, Weissenborn, Krithara, Petridis, Polychronopoulos,
  Almirantis, Pavlopoulos, Baskiotis, Gallinari, Arti{\`{e}}res, Ngomo, Heino,
  Gaussier, Barrio{-}Alvers, Schroeder, Androutsopoulos, and
  Paliouras}]{tsatsaronis2015bioasq}
George Tsatsaronis, Georgios Balikas, Prodromos Malakasiotis, Ioannis Partalas,
  Matthias Zschunke, Michael~R. Alvers, Dirk Weissenborn, Anastasia Krithara,
  Sergios Petridis, Dimitris Polychronopoulos, Yannis Almirantis, John
  Pavlopoulos, Nicolas Baskiotis, Patrick Gallinari, Thierry Arti{\`{e}}res,
  Axel{-}Cyrille~Ngonga Ngomo, Norman Heino, {\'{E}}ric Gaussier, Liliana
  Barrio{-}Alvers, Michael Schroeder, Ion Androutsopoulos, and Georgios
  Paliouras. 2015.
\newblock \href {https://doi.org/10.1186/s12859-015-0564-6} {An overview of the
  {BIOASQ} large-scale biomedical semantic indexing and question answering
  competition}.
\newblock \emph{{BMC} Bioinform.}, 16:138:1--138:28.

\bibitem[{Vilalta and Drissi(2002)}]{vilalta2002metalearn}
Ricardo Vilalta and Youssef Drissi. 2002.
\newblock \href {https://doi.org/10.1023/A:1019956318069} {A perspective view
  and survey of meta-learning}.
\newblock \emph{Artif. Intell. Rev.}, 18(2):77--95.

\bibitem[{Voorhees and Tice(2000)}]{voorhees2000trec}
Ellen~M. Voorhees and Dawn~M. Tice. 2000.
\newblock \href {https://doi.org/10.1145/345508.345577} {Building a question
  answering test collection}.
\newblock In \emph{Proceedings of the 23rd Annual International ACM SIGIR
  Conference on Research and Development in Information Retrieval}, SIGIR '00,
  page 200–207, New York, NY, USA. Association for Computing Machinery.

\bibitem[{Wang et~al.(2018)Wang, Singh, Michael, Hill, Levy, and
  Bowman}]{wang2018glue}
Alex Wang, Amanpreet Singh, Julian Michael, Felix Hill, Omer Levy, and Samuel
  Bowman. 2018.
\newblock \href {https://doi.org/10.18653/v1/W18-5446} {{GLUE}: A multi-task
  benchmark and analysis platform for natural language understanding}.
\newblock In \emph{Proceedings of the 2018 {EMNLP} Workshop {B}lackbox{NLP}:
  Analyzing and Interpreting Neural Networks for {NLP}}, pages 353--355,
  Brussels, Belgium. Association for Computational Linguistics.

\bibitem[{Wei et~al.(2022)Wei, Bosma, Zhao, Guu, Yu, Lester, Du, Dai, and
  Le}]{wei2022zeroshot}
Jason Wei, Maarten~Paul Bosma, Vincent Zhao, Kelvin Guu, Adams~Wei Yu, Brian
  Lester, Nan Du, Andrew~Mingbo Dai, and Quoc~V. Le. 2022.
\newblock \href {https://openreview.net/forum?id=gEZrGCozdqR} {Finetuned
  language models are zero-shot learners}.
\newblock In \emph{International Conference on Learning Representations}.

\bibitem[{Williams et~al.(2018)Williams, Nangia, and Bowman}]{williams2018mnli}
Adina Williams, Nikita Nangia, and Samuel Bowman. 2018.
\newblock \href {https://doi.org/10.18653/v1/N18-1101} {A broad-coverage
  challenge corpus for sentence understanding through inference}.
\newblock In \emph{Proceedings of the 2018 Conference of the North {A}merican
  Chapter of the Association for Computational Linguistics: Human Language
  Technologies, Volume 1 (Long Papers)}, pages 1112--1122, New Orleans,
  Louisiana. Association for Computational Linguistics.

\bibitem[{Xie et~al.(2022)Xie, Raghunathan, Liang, and Ma}]{xie2022bayesianicl}
Sang~Michael Xie, Aditi Raghunathan, Percy Liang, and Tengyu Ma. 2022.
\newblock \href {https://openreview.net/forum?id=RdJVFCHjUMI} {An explanation
  of in-context learning as implicit bayesian inference}.
\newblock In \emph{International Conference on Learning Representations}.

\bibitem[{Ye et~al.(2021)Ye, Lin, and Ren}]{ye2021crossfit}
Qinyuan Ye, Bill~Yuchen Lin, and Xiang Ren. 2021.
\newblock \href {https://doi.org/10.18653/v1/2021.emnlp-main.572}
  {{C}ross{F}it: A few-shot learning challenge for cross-task generalization in
  {NLP}}.
\newblock In \emph{Proceedings of the 2021 Conference on Empirical Methods in
  Natural Language Processing}, pages 7163--7189, Online and Punta Cana,
  Dominican Republic. Association for Computational Linguistics.

\bibitem[{Yu et~al.(2018)Yu, Guo, Yi, Chang, Potdar, Cheng, Tesauro, Wang, and
  Zhou}]{yu2018diverse}
Mo~Yu, Xiaoxiao Guo, Jinfeng Yi, Shiyu Chang, Saloni Potdar, Yu~Cheng, Gerald
  Tesauro, Haoyu Wang, and Bowen Zhou. 2018.
\newblock \href {https://doi.org/10.18653/v1/N18-1109} {Diverse few-shot text
  classification with multiple metrics}.
\newblock In \emph{Proceedings of the 2018 Conference of the North {A}merican
  Chapter of the Association for Computational Linguistics: Human Language
  Technologies, Volume 1 (Long Papers)}, pages 1206--1215, New Orleans,
  Louisiana. Association for Computational Linguistics.

\bibitem[{Zhong et~al.(2021)Zhong, Lee, Zhang, and Klein}]{zhong2021adapting}
Ruiqi Zhong, Kristy Lee, Zheng Zhang, and Dan Klein. 2021.
\newblock \href {https://doi.org/10.18653/v1/2021.findings-emnlp.244} {Adapting
  language models for zero-shot learning by meta-tuning on dataset and prompt
  collections}.
\newblock In \emph{Findings of the Association for Computational Linguistics:
  EMNLP 2021}, pages 2856--2878, Punta Cana, Dominican Republic. Association
  for Computational Linguistics.

\end{thebibliography}
\bibliographystyle{acl_natbib}

\appendix

\section{Tasks}\label{app:tasks}
\paragraph{Meta-training.} Our meta-training data is from MetaICL's \citep{min2022metaicl} meta-training sets. Specifically, we use the QA task collection from the paper, which is a mixture of \textsc{CrossFit} and \textsc{UnifiedQA} tasks as shown in Table \ref{tab:metatrain_tasks}. We exclude any task on which we evaluate. As in MetaICL, we subsample 16,384 examples per task such that no individual task is overrepresented during meta-training. Some tasks are sampled twice due to the inclusion of both \textsc{CrossFit} and \textsc{UnifiedQA} versions of some tasks, as in \citet{min2022metaicl}.
\begin{table}[h]
    \centering
    \resizebox{\linewidth}{!}{
    \begin{tabular}{p{1.4\linewidth}}
    \toprule
    All meta-training tasks: \\
    biomrc, boolq, freebase\_qa, hotpot\_qa, kilt\_hotpotqa, kilt\_nq, kilt\_trex, kilt\_zsre, lama-conceptnet, lama-google\_re, lama-trex, mc\_taco, numer\_sense, quoref, ropes, search\_qa, superglue-multirc, superglue-record, tweet\_qa, web\_questions, unifiedqa:boolq, unifiedqa:commonsenseqa, unifiedqa:drop, unifiedqa:narrativeqa, unifiedqa:natural\_questions\_with\_dpr\_para,  unifiedqa:newsqa, unifiedqa:physical\_iqa, unifiedqa:quoref, unifiedqa:race\_string, unifiedqa:ropes, unifiedqa:social\_iqa, unifiedqa:winogrande\_xl \\ 
    \midrule
    Subsampled by similarity: \\
    biomrc, boolq, freebase\_qa, hotpot\_qa, lama-google\_re, quoref, ropes, superglue-multirc, superglue-record, unifiedqa:boolq,  unifiedqa:commonsenseqa, unifiedqa:drop, unifiedqa:narrativeqa, unifiedqa:natural\_questions\_with\_dpr\_para, unifiedqa:newsqa, unifiedqa:quoref, unifiedqa:race\_string, unifiedqa:ropes \\
    \bottomrule
    \end{tabular}}
    \caption{Tasks used in our meta-training data. We subsample 16,384 examples per task to ensure balanced supervision during meta-training. All tasks are from \textsc{CrossFit} unless prefixed with ``unifiedqa:''.}
    \label{tab:metatrain_tasks}
\end{table}

We also perform a targeted subsampling procedure, where we select tasks by semantic similarity to our evaluation tasks. For this, we compute the mean pairwise semantic similarity between a meta-training task's examples and one 16-shot split of each of our evaluation tasks, then select meta-training tasks in decreasing order of similarity. Semantic similarity is computed by calculating the cosine similarity of the sentence embeddings from the \texttt{stsb-roberta-base-v2} model in SentenceTransformers \citep{reimers2019sentence}.

\paragraph{Demonstrations.}
\begin{table}[t]
    \centering
    \resizebox{\linewidth}{!}{
    \begin{tabular}{p{1.4\linewidth}}
    \toprule
    Demonstration task bank: \\
    unifiedqa:ai2\_science\_middle, unifiedqa:boolq, unifiedqa:commonsenseqa, unifiedqa:drop, unifiedqa:mctest, unifiedqa:narrativeqa, unifiedqa:natural\_questions\_with\_dpr\_para,  unifiedqa:newsqa, unifiedqa:openbookqa, unifiedqa:openbookqa\_with\_ir, unifiedqa:physical\_iqa, unifiedqa:quoref, unifiedqa:race\_string, unifiedqa:ropes, unifiedqa:social\_iqa, unifiedqa:winogrande\_xl \\ 
    \bottomrule
    \end{tabular}}
    \caption{Tasks used in our demonstration memory bank. Note that there is no subsampling within each task, since the retriever can simply ignore irrelevant demonstrations. All tasks are from \textsc{UnifiedQA}.}
    \label{tab:demo_tasks}
\end{table}

Our demonstrations are from the \textsc{UnifiedQA} collection, which includes extractive, abstractive, and multiple-choice QA tasks as shown in Table \ref{tab:demo_tasks}. We exclude any task on which we evaluate.

Note that there is some overlap between the demonstration set and the meta-training set, though the demonstrations contain the correct answer whereas the meta-training examples do not.

\section{Format Tuning for Multiple-choice QA}\label{app:qasc_formats}
\citet{chada2021fewshotqa} observe significant performance gains by simply changing the format of the QA inputs and outputs. We use a format similar to theirs for most QA tasks, but it is not immediately clear how to extend the question/answer/context format to multiple-choice QA, or if including the answer options in the context would be helpful at all. Thus, we try three different formats for QASC and compare performance.

\begin{table*}[]
    \centering
    \resizebox{\linewidth}{!}{
    \begin{tabular}{p{3cm}p{5cm}p{5cm}r}
    \toprule
    Format name & Format & Example & $F_1$ \\
    \midrule
    Options in question, generate letter & question: $q$? $\{a_A,\ldots,a_H\}$ \texttt{\textbackslash{}n} answer: \texttt{[MASK]} \texttt{\textbackslash{}n} context: $c_1$. $c_2$. $\Rightarrow$ question: $q$? \texttt{\textbackslash{}n} answer: $i$ & question: What does sunlight do for a plant? (A) during the day (B) Kills it (C) it can be seen (D) Helps it survive (E) Helps it drink water (F) It gets heated up (G) adding heat (H) Makes the color darker \texttt{\textbackslash{}n} answer: \texttt{[MASK]} \texttt{\textbackslash{}n} context: A plant requires food for survival. All plants require sunlight to make their food. $\Rightarrow$ question: $\ldots$ \texttt{\textbackslash{}n} answer: D & 15.6 \\
    Options in question, generate answer & question: $q$? $\{a_A,\ldots,a_H\}$ \texttt{\textbackslash{}n} answer: \texttt{[MASK]} \texttt{\textbackslash{}n} context: $c_1$. $c_2$. $\Rightarrow$ question: $q$? \texttt{\textbackslash{}n} answer: $a$ & question: What does sunlight do for a plant? (A) during the day (B) Kills it (C) it can be seen (D) Helps it survive (E) Helps it drink water (F) It gets heated up (G) adding heat (H) Makes the color darker \texttt{\textbackslash{}n} answer: \texttt{[MASK]} \texttt{\textbackslash{}n} context: A plant requires food for survival. All plants require sunlight to make their food. $\Rightarrow$ question: $\ldots$ \texttt{\textbackslash{}n} answer: Helps it survive & 39.4 \\
    Options in context, generate answer & question: $q$? \texttt{\textbackslash{}n} answer: \texttt{[MASK]} \texttt{\textbackslash{}n} context: $\{a_A,\ldots,a_H\}$. $c_1$. $c_2$. $\Rightarrow$ question: $q$? \texttt{\textbackslash{}n} answer: $a$ & question: What does sunlight do for a plant? \texttt{\textbackslash{}n} answer: \texttt{[MASK]} \texttt{\textbackslash{}n} context: (A) during the day (B) Kills it (C) it can be seen (D) Helps it survive (E) Helps it drink water (F) It gets heated up (G) adding heat (H) Makes the color darker. A plant requires food for survival. All plants require sunlight to make their food. $\Rightarrow$ question: $\ldots$ \texttt{\textbackslash{}n} answer: Helps it survive & \textbf{82.6} \\
    No options, generate answer & question: $q$? \texttt{\textbackslash{}n} answer: \texttt{[MASK]} \texttt{\textbackslash{}n} context: $c_1$. $c_2$. $\Rightarrow$ question: $q$? \texttt{\textbackslash{}n} answer: $a$ & question: What does sunlight do for a plant? \texttt{\textbackslash{}n} answer: \texttt{[MASK]} \texttt{\textbackslash{}n} context: A plant requires food for survival. All plants require sunlight to make their food. $\Rightarrow$ question: $\ldots$ \texttt{\textbackslash{}n} answer: Helps it survive & 49.8 \\
    \bottomrule
    \end{tabular}}
    \caption{The formats we try for QASC and 16-shot $F_1$ scores from BART\textsubscript{large} (no retrieval) after fine-tuning on each format. We find that generating the answer is better than just generating the letter label, that including the options in the context is helpful, and that excluding the options from the context is harmful to performance. ``$\Rightarrow$'' separates the input from the output sequence, and ``\texttt{\textbackslash{}n}'' indicates a newline.}
    \label{tab:qasc_formats}
\end{table*}

Every example consists of a question $q$, two context sentences $c_1$ and $c_2$, a set of 8 answer options with letter labels \{$a_A$,~$a_B$,~$\ldots$,~$a_H$\}, and a correct answer $a\in\{a_A,\ldots,a_H\}$. We can generate either the full answer string, or the letter label of the answer $i$, where $i\in\{A,B,\ldots,H\}$. We try putting the answer options in the question or the context, excluding the answer options altogether, generating the answer string $a$, and generating the answer letter $i$.

Our results using BART\textsubscript{large} (Table~\ref{tab:qasc_formats}) indicate that generating the answer is better than just generating the letter label, that including the options in the context is helpful, and that excluding the options from the context or putting the options in the question is harmful to performance. The performance gap between different formats is \emph{very} large, which aligns with the findings of \citet{chada2021fewshotqa}: using an example format aligned with the model's pre-training format is one of the most important factors contributing to few-shot performance.
\end{document}